\pdfoutput=1
\documentclass[11pt]{article}
\usepackage[]{acl}
\usepackage{times}
\usepackage{latexsym}
\usepackage{url}
\usepackage{tikz}
\usepackage{tikz-qtree}
\usepackage{subfigure}
\usepackage{pgfplots}
\pgfplotsset{compat=1.8}
\usepgfplotslibrary{statistics}
\usetikzlibrary{patterns}
\usepgfplotslibrary{groupplots}
\usepackage{booktabs}
\usepackage{graphicx}
\usepackage{multirow}
\usepackage{caption}
\usepackage{stackengine}
\usepackage[T1]{fontenc}
\usepackage[utf8]{inputenc}
\usepackage{microtype}

\title{Towards Understanding Large-Scale Discourse Structures in Pre-Trained and Fine-Tuned Language Models}

\author{Patrick Huber and Giuseppe Carenini\\
  Department of Computer Science \\
  University of British Columbia \\
  Vancouver, BC, Canada, V6T 1Z4 \\
  {\tt \{huberpat, carenini\}@cs.ubc.ca}}

\begin{document}
\maketitle
\begin{abstract}
With a growing number of BERTology work analyzing different components of pre-trained language models, we extend this line of research through an in-depth analysis of discourse information in pre-trained and fine-tuned language models. We move beyond prior work along three dimensions: First, we describe a novel approach to infer discourse structures from arbitrarily long documents. Second, we propose a new type of analysis to explore where and how accurately intrinsic discourse is captured in the BERT and BART models. Finally, we assess how similar the generated structures are to a variety of baselines as well as their distribution within and between models.
\end{abstract}

\section{Introduction}
Transformer-based machine learning models are an integral part of many recent improvements in Natural Language Processing (NLP). With their rise spearheaded by \citet{vaswani2017attention}, the pre-training/fine-tuning paradigm has gradually replaced previous approaches based on architecture engineering, with transformer models such as BERT \cite{devlin2018bert}, BART \cite{lewis2020bart}, RoBERTa \cite{liu2019roberta} and others delivering state-of-the-art performances on a wide variety of tasks. Besides their strong empirical results on most real-world problems, such as summarization \cite{zhang2020pegasus, xiao2021primer}, question-answering \cite{joshi2020spanbert, ouguz2021domain} and sentiment analysis \cite{adhikari2019docbert, yang2019xlnet}, uncovering what kind of linguistic knowledge is captured by this new type of pre-trained language models (PLMs) has become a prominent question by itself. As part of this line of research, called \textit{BERTology} \cite{rogers2020primer}, researchers explore the amount of linguistic understanding encapsulated in PLMs, exposed through either external probing tasks \cite{raganato2018analysis, zhu2020examining, koto2021discourse} or unsupervised methods \cite{wu2020perturbed, pandia2021pragmatic} to analyze the syntactic structures (e.g., \citet{hewitt-manning-2019-structural, 
wu2020perturbed}), relations \cite{papanikolaou-etal-2019-deep}, ontologies \cite{michael-etal-2020-asking} and, to a more limited extend, discourse related behaviour \cite{zhu2020examining, koto2021discourse, pandia2021pragmatic}. 

Generally speaking, while most previous \textit{BERTology} work has focused on either sentence level phenomena or connections between adjacent sentences, large-scale semantic and pragmatic structures (oftentimes represented as discourse trees/graphs) have been less explored. These structures (e.g., discourse trees) play a fundamental role in expressing the intent of multi-sentential documents and, not surprisingly, have been shown to benefit many NLP tasks such as summarization \cite{gerani2019modeling}, sentiment analysis \cite{bhatia2015better, nejat2017exploring, hogenboom2015using} and text classification \cite{ji2017neural}. 

With multiple different theories for discourse proposed in the past, the RST \cite{mann1988rhetorical} and PDTB \cite{prasadpenn} frameworks have received most attention. 
RST-style discourse structures thereby consist of a single rooted tree covering whole documents, comprising of: (1) A tree structure, combining clause-like sentence fragments (Elementary Discourse Units, short: EDUs) into a discourse constituency tree, (2) Nuclearity, assigning every tree-branch primary (\textit{Nucleus}) or peripheral (\textit{Satellite}) importance in a local context and (3) Relations, defining the connection and direction between siblings in the tree. 
 
Given the importance of large-scale discourse structures, we extend the line of \textit{BERTology} research with novel experiments to test for the presence of intrinsic discourse information in established PLMs. More specifically, we aim to better understand to what extend RST-style discourse information is stored as latent trees in encoder self-attention matrices\footnote{We focus on discourse structure and nuclearity, leaving the relation classification for future work.}. While we focus on the RST formalism in this work, our presented methods are theory-agnostic and, hence, applicable to discourse structures in a broader sense, including other tree-based theories, such as PDTB. Our contributions in this paper are:\\
\textbf{(1)} A novel approach to extract discourse information from arbitrarily long documents with limited-size transformer models. This is a non-trivial issue, which has been mostly by-passed in previous work through the use of proxy tasks.\\
\textbf{(2)} An exploration of discourse information locality across pre-trained and fine-tuned language models, finding that discourse is consistently captured in a fixed subset of self-attention heads. \\
\textbf{(3)} An in-depth analysis of the discourse quality in pre-trained language models and their fine-tuned extensions. We compare constituency and dependency structures of 2 PLMs fine-tuned on 4 tasks and 7 fine-tuning datasets to gold-standard discourse trees, finding that the captured discourse structures outperform simple baselines by a wide margin and even show superior performance compared to distantly supervised models.\\
\textbf{(4)} A similarity analysis between PLM inferred discourse trees and supervised, distantly supervised and simple baselines, which reveals that PLM constituency discourse trees do align relatively well with previously proposed supervised models, but also capture complementary information, making them a valuable resource for ensemble methods. \\
\textbf{(5)} A detailed look at information redundancy in self-attention heads to better understand the structural overlap between self-attention matrices and models. Our results indicate that similar discourse information is consistently captured in the same heads, even across fine-tuning tasks.

\section{Related Work}
At the base of our work are two of the most popular and frequently used PLMs: BERT \cite{devlin2018bert} and BART \cite{lewis2020bart}. We choose these two popular approaches in our study due to their complementary nature (encoder-only vs. encoder-decoder) and based on previous work by \citet{zhu2020examining} and \citet{koto2021discourse}, showing the effectiveness of BERT and BART models for discourse related tasks.

Our work is further related to the field of discourse parsing. With a rich history of traditional machine learning models (e.g., \citet{hernault2010hilda, ji2014representation, joty2015codra, wang2017two}\textit{, inter alia}), recent approaches slowly shifted to successfully incorporate a variety of PLMs into the process of discourse prediction, such as ELMo embeddings \cite{kobayashi2019split},  XLNet \cite{nguyen2021rst}, BERT \cite{koto2021top}, RoBERTa \cite{guz2020unleashing} and SpanBERT \cite{guz2020coreference}. Despite these works showing the usefulness of PLMs for discourse parsing, all of them cast the task into a ``local" problem,  using only partial information through the shift-reduce framework \cite{guz2020unleashing, guz2020coreference}, natural document breaks (e.g. paragraphs \cite{kobayashi2020top}) or by framing the task as an inter-EDU sequence labelling problem on partial documents \cite{koto2021top}. However, since we believe that the true benefit of discourse information only emerges when complete documents are considered, we propose a new approach to connect PLMs and discourse structures in a ``global'' manner, superseding the local proxy-tasks with a new methodology to explore arbitrarily long documents.

Aiming to better understand what information is captured in PLMs, the line of \textit{BERTology} research has recently emerged \cite{rogers2020primer}, with early work mostly focusing on the syntactic capacity of PLMs  \cite{hewitt-manning-2019-structural, jawahar-etal-2019-bert, kim2019pre}, in parts also exploring the internal workings of transformer-based models (e.g., self-attention matrices \cite{raganato2018analysis, marevcek2019balustrades}). More recent work started to explore the alignment of PLMs with discourse information, encoding semantic and pragmatic knowledge. Along those lines, \citet{wu2020perturbed} present a parameter-free probing task for both, syntax and discourse. Compared to our work, their tree inference approach is however computationally expensive and only explores the outputs of the BERT model. Further, \citet{zhu2020examining} use $24$ hand-crafted rhetorical features to execute three different supervised probing tasks, showing promising performance of the BERT model. Similarly, \citet{pandia2021pragmatic} aim to infer pragmatics through the prediction of discourse connectives by analyzing the model inputs and outputs and \citet{koto2021discourse} analyze discourse in PLMs through seven supervised probing tasks, finding that BART and BERT contain most information related to discourse. In contrast to the approach taken by both \citet{zhu2020examining} and \citet{koto2021discourse}, we use an unsupervised methodology to test the amount of discourse information stored in PLMs (which can also conveniently be used to infer discourse structures for new and unseen documents) and extend the work by \citet{pandia2021pragmatic} by taking a closer look at the internal workings of the self-attention component.

\begin{figure}[t]
    \centering
    \setlength{\belowcaptionskip}{-15pt}
    \includegraphics[width=.8\linewidth]{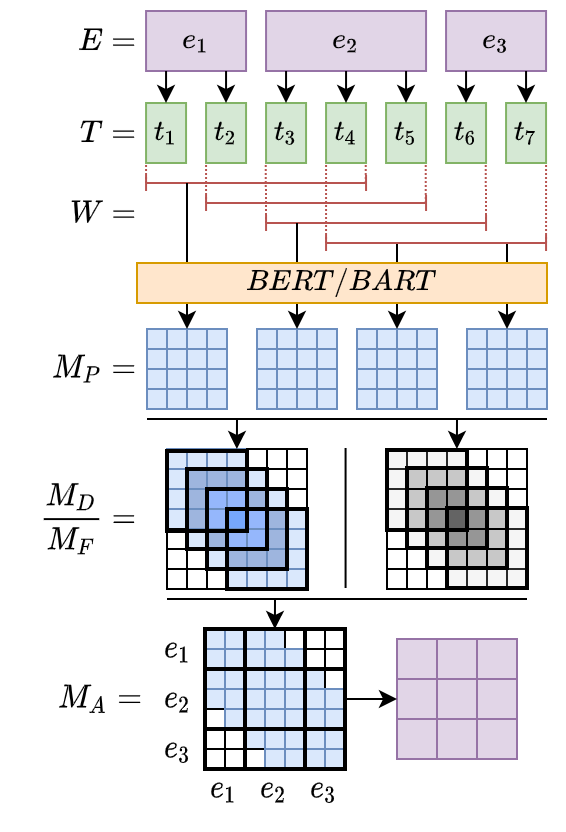}
    \caption{Small-scale example of the discourse extraction approach. Purple=EDUs, green=sub-word embeddings, red=input slices of size $t_{max}$, orange=PLM, blue=self-attention values, grey-scale=frequency count.}
    \label{fig:disc_ext}
\end{figure}

Looking at all these prior works analyzing the amount of discourse in PLMs, structures are solely explored through the use of proxy tasks, such as connective prediction \cite{pandia2021pragmatic}, relation classification \cite{kurfali2021probing} and others \cite{koto2021discourse}. However, despite the difficulties of encoding arbitrarily long documents, we believe that to systematically explore the relationship between PLMs and discourse, considering complete documents is imperative. 
Along these lines, recent work started to tackle the inherent input-length limitation of general transformer models through additional recurrence \cite{dai2019transformer}, compression modules \cite{rae2019compressive} or sparse patterns (e.g., \citet{kitaev2020reformer, beltagy2020longformer}). 
Still mostly based on established PLMs (e.g., BERT) and with no dominant solution yet, we believe that even with the input length restriction being actively tackled, an in-depth analysis of traditional PLMs with discourse is highly valuable to establish a solid understanding of intrinsic linguistic properties.

Besides the described \textit{BERTology} work, we got encouraged to explore fine-tuned extensions of standard PLMs through previous work showing the benefit of discourse parsing for many downstream tasks, such as summarization \cite{gerani2019modeling}, sentiment analysis \cite{bhatia2015better, nejat2017exploring, hogenboom2015using} and text classification \cite{ji2017neural}. Conversely, recent work also shows promising results when inferring discourse structures from related downstream tasks, such as sentiment analysis \cite{huber2020mega} and summarization \cite{xiao2021predicting}. Given this bidirectional synergy, we move beyond traditional experiments focusing on standard PLMs and additionally explore discourse structures of fine-tuned PLMs.

\section{Discourse Extraction Method}
\label{sec:extract}
With PLMs rather well analyzed according to their syntactic capabilities, large-scale discourse structures have been less explored. One reason for this is the input length constraint of transformer models. While this is generally not prohibitive for intra-sentence syntactic structures (e.g., presented in \citet{wu2020perturbed}), it does heavily influence large-scale discourse structures, operating on complete (potentially long) documents. Overcoming this limitation is non-trivial, since traditional transformer-based models only allow for fixed, short inputs. Aiming to systematically explore the ability of PLMs to capture discourse, we investigate a novel way to effectively extract discourse structures from the self-attention component of the BERT and BART models. We thereby extend the tree-generation approach proposed in \citet{xiao2021predicting} to support the input length constraints of standard PLMs using a sliding-window approach in combination with matrix frequency normalization and an EDU aggregation method. 

\paragraph{The Tree Generation Procedure} by \citet{xiao2021predicting} explores a two-stage approach to obtain discourse structures from a transformer model, by-passing the input-length constraint. Using the intuition that the self-attention score between any two EDUs is an indicator of their semantic/pragmatic relatedness and hence should influence their distance in a projective discourse tree, they use the CKY dynamic programming approach \cite{jurafsky2014speech} to generate constituency trees based on the internal self-attention of the transformer model. To generate dependency trees, a similar intuition is applied when inferring discourse trees using the Eisner \cite{eisner-1996-three} algorithm. 
Since we explore the discourse information captured in standard PLMs, we cannot directly transfer the two-stage approach in \citet{xiao2021predicting}\footnote{For more information on the tree-generation approach, we refer interested readers to \citet{xiao2021predicting}.}. Instead, we propose a new method to overcome the length-limitation of the transformer model.

\paragraph{The Sliding-Window Approach} is at the core of our new methodology to overcome the input-length constraint. We first tokenize arbitrarily long documents with $n$ EDUs $E = \{e_1, ..., e_n\}$ into the respective sequence of $m$ sub-word tokens $T = \{t_1, ... t_m\}$ with $n \ll m$, according to the PLM tokenization method (WordPiece for BERT, Byte-Pair-Encoding for BART). Using the sliding window approach, we subdivide the $m$ sub-word tokens into sequences of maximum input length $t_{max}$, defined by the PLM. Using a stride of $1$, we generate $(m-t_{max})+1$ sliding windows $W$, feed them into the PLM, and extract the resulting $t_{max} \times t_{max}$ partial self-attention matrices $M_{P}$ for a specific self-attention head\footnote{We omit the self-attention indexes for better readability.}. 

\paragraph{The Frequency Normalization Method} allows us to combine the partially overlapping self-attention matrices $M_{P}$ into a single document-level matrix $M_{D}$ of size $m \times m$. To this end, we interpolate multiple overlapping windows by adding up the self-attention cells of tokens $t_i$, while keeping track of the number of overlaps in a separate $m \times m$ frequency matrix $M_{F}$. We then divide $M_D$ by the frequency matrix $M_F$, to generate a frequency normalized self-attention matrix.
 
\paragraph{The EDU Aggregation} is the final processing step to obtain the document-level self-attention matrix $M_A$. In this step, the $m$ sub-word tokens $T = \{t_1, ... t_m\}$ are aggregated back into $n$ EDUs $E = \{e_1, ..., e_n\}$ by computing the average bidirectional self-attention score between any two EDUs in $\frac{M_D}{M_F}$. Then, we use the resulting $n \times n$ matrix $M_A$ as the input to the CKY/Eisner discourse tree generation methods. Figure \ref{fig:disc_ext} visualizes the complete process on a small scale example. 
\section{Experimental Setup}

\subsection{Pre-Trained Models}
We select the \textit{BERT-base} ($110$ million parameters) and \textit{BART-large} ($406$ million parameters) models for our experiments. We choose these models for their diverse objectives (encoder-only vs. encoder-decoder), popularity for diverse fine-tuning tasks, and their prior exploration in regards to discourse \cite{zhu2020examining,koto2021discourse}. For the \textit{BART-large} model, we limit our analysis to the encoder, as motivated in \citet{koto2021discourse}, leaving experiments with the decoder for future work.

\subsection{Fine-Tuning Tasks and Datasets}
We explore the BERT model fine-tuned on two classification tasks, namely sentiment analysis and natural language inference (NLI). For our analysis on BART, we select the abstractive summarization and question answering tasks. Table \ref{tab:datasets} summarizes the 7 datasets used to fine-tune PLMs in this work, along with their underlying tasks and domains\footnote{
We exclusively analyze published models provided on the huggingface platform, further specified in Appendix \ref{app:links}.}.

\begin{table}[t]
    \centering
    \setlength{\belowcaptionskip}{-5pt}
    \resizebox{\linewidth}{!}{
    \begin{tabular}{llll}
        \toprule
        Dataset & Task & Domain \\
        \midrule
        IMDB\shortcite{diao2014jointly} & Sentiment & Movie Reviews \\
        Yelp\shortcite{zhang2015character} & Sentiment & Reviews  \\
        SST-2\shortcite{socher-etal-2013-recursive} & Sentiment & Movie Reviews \\
        MNLI\shortcite{N18-1101} & NLI & Range of Genres \\
        CNN-DM\shortcite{nallapati-etal-2016-abstractive} & Summarization & News \\
        XSUM\shortcite{Narayan2018DontGM} & Summarization & News \\
        SQuAD\shortcite{rajpurkar2016squad} & Question-Answering & Wikipedia \\
        \bottomrule
    \end{tabular}}
    \caption{The seven fine-tuning datasets used in this work along with the underlying tasks and domains.}
    \label{tab:datasets}
\end{table}

\subsection{Evaluation Treebanks}
\textbf{RST-DT} \cite{carlson2002rst} is the largest English RST-style discourse treebank, containing $385$ Wall-Street-Journal articles, annotated with full constituency discourse trees. To generate additional dependency trees, we apply the conversion algorithm proposed in \citet{li-etal-2014-text}. \\
\textbf{GUM} \cite{Zeldes2017} is a steadily growing treebank of richly annotated texts. In the current version $7.3$, the dataset contains $168$ documents from $12$ genres, annotated with full RST-style constituency and dependency discourse trees. 

All evaluations shown in this paper are executed on the $38$ and $20$ documents in the RST-DT and GUM test-sets, to be comparable with previous baselines and supervised models. 

\begin{figure}[t]
    \centering
    \setlength{\belowcaptionskip}{-5pt}
    \subfigure[BERT: PLM, +IMDB, +Yelp, +SST-2, +MNLI]{
        \includegraphics[width=.85\linewidth]{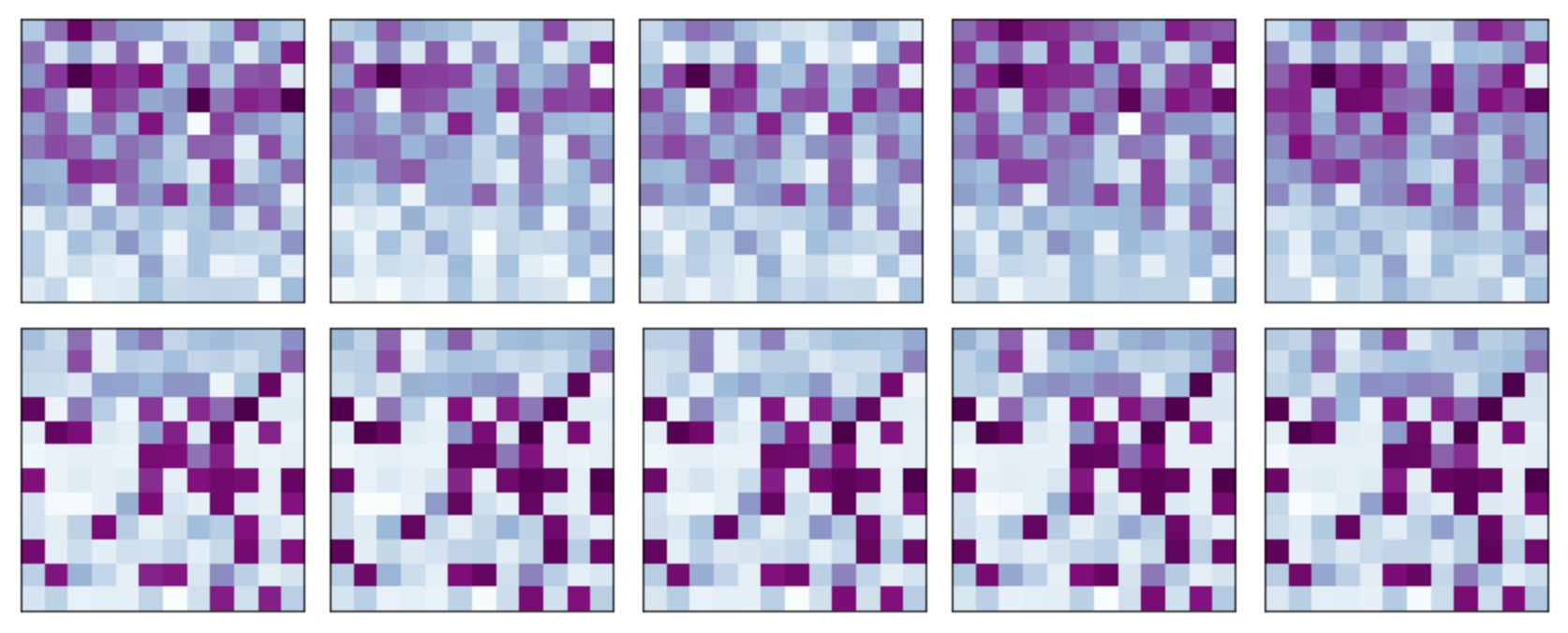}}
    \centering
    \subfigure[BART: PLM, +CNN-DM, +XSUM, +SQuAD]{
    \includegraphics[width=.85\linewidth]{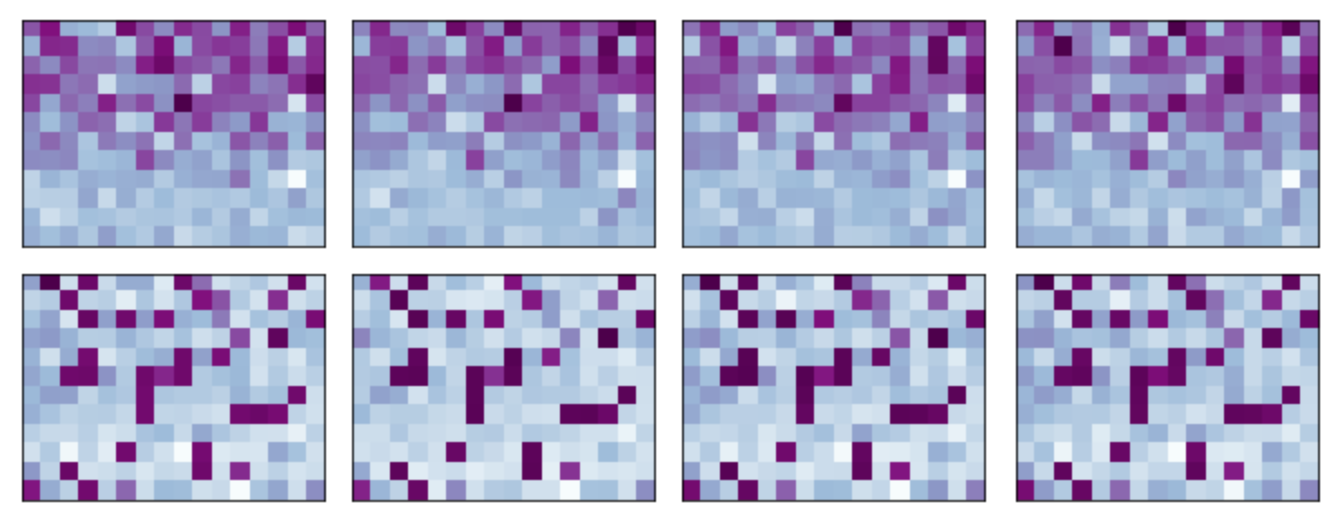}}
    \caption{Constituency (top) and dependency (bottom) discourse tree evaluation of BERT (a) and BART (b) models on GUM. Purple=high score, Blue=low score. Heads presented left-to-right, high layers on top. \\+ indicates fine-tuning dataset.}
    \label{fig:gum_heatmap_collection}
\end{figure}

\subsection{Baselines and Evaluation Metrics}
\label{sec:baselines}
\textbf{Simple Baselines:} We compare the inferred constituency trees against right- and left-branching structures. For dependency trees, we evaluate against simple chain and inverse chain structures. \\
\textbf{Distantly Supervised Baselines:} We compare our results against the approach by \citet{xiao2021predicting}, using similar CKY and Eisner tree-generation methods to infer constituency and dependency tree structures from their summarization model trained on the CNN-DM and New York Times (NYT) corpora (called \textit{Sum\textsubscript{CNN-DM}} and \textit{Sum\textsubscript{NYT}})\footnote{\url{www.github.com/Wendy-Xiao/summ_guided_disco_parser}}.\\
\textbf{Supervised Baseline:} We select the popular Two-Stage discourse parser \cite{wang2017two} as our supervised baseline, due to its strong performance, available model checkpoints and code\footnote{\url{www.github.com/yizhongw/StageDP}}, as well as the traditional architecture. We use the published Two-Stage parser checkpoint on RST-DT (from here on called \textit{Two-Stage\textsubscript{RST-DT}}) and re-train the discourse parser on GUM (\textit{Two-Stage\textsubscript{GUM}}).
We convert the generated constituency structures into dependency trees following \citet{li-etal-2014-text}.\\
\textbf{Evaluation Metrics:} 
We apply the original parseval score to compare discourse constituency structures with gold-standard treebanks, as argued in \citet{morey2017much}. To evaluate the generated dependency structures, we use the Unlabeled Attachment Score (UAS).

\section{Experimental Results}
\subsection{Discourse Locality}
\label{sec:PLM_and_extensions}
Our discourse tree generation approach described in section \ref{sec:extract} directly uses self-attention matrices to generate discourse trees. The standard BERT model contains $144$ of those self-attention matrices ($12$ layers, $12$ self-attention heads each), all of which potentially encode discourse structures. For the BART model, this number is even higher, consisting of $12$ layers with $16$ self-attention heads each. With prior work suggesting the locality of discourse information in PLMs (e.g., \citet{raganato2018analysis, marevcek2019balustrades,xiao2021predicting}), we analyze every self-attention matrix individually to gain a better understanding of their alignment with discourse information. 

Besides investigating standard PLMs, we also explore the robustness of discourse information across fine-tuning tasks. We believe that this is an important step to better understand if the captured discourse information is general and robust, or if it is ``re-learned'' from scratch for downstream tasks. To the best of our knowledge, no previous analysis of this kind has been performed in the literature.

To this end, Figure \ref{fig:gum_heatmap_collection} shows the constituency and dependency structure overlap of the generated discourse trees from 
every individual self-attention head with the gold-standard tree structures of the GUM dataset\footnote{The analysis on RST-DT shows similar trends and can be found in Appendix \ref{sec:app_self_attn_test}.}. The heatmaps clearly show that constituency discourse structures are mostly captured in higher layers, while dependency structures are scattered throughout. Comparing the patterns between models, we find that, despite being fine-tuned on different downstream tasks, the discourse information is consistently encoded in the same self-attention heads. Even though the best performing self-attention matrix is not consistent, discourse information is clearly captured in a ``local" subset of self-attention heads across all presented fine-tuning task. 
This plausibly suggests that the discourse information in pre-trained BERT and BART models is robust and general, requiring only minor adjustments depending on the fine-tuning task.

\begin{table}[t]
    \centering
    \setlength{\belowcaptionskip}{-15pt}
    \resizebox{\linewidth}{!}{
    \begin{tabular}{lrrrr}
        \toprule
        \multirow{2}{*}{Model} & \multicolumn{2}{c}{RST-DT} & \multicolumn{2}{c}{GUM} \\
         & Span & UAS & Span & UAS \\
        \midrule
        \multicolumn{5}{c}{BERT} \\
        \midrule
        rand. init & $\downarrow$ 25.5 & $\downarrow$ 13.3 & $\downarrow$ 23.2 & $\downarrow$ 12.4 \\
        PLM & $\bullet$ 35.7 & $\bullet$ 45.3 & $\bullet$ 33.0 & $\bullet$ 45.2 \\
        ~~+ IMDB & $\downarrow$ 35.4 & $\downarrow$ 42.8 & $\bullet$ 33.0 & $\downarrow$ 43.3 \\
        ~~+ Yelp & $\downarrow$ 34.7 & $\downarrow$ 42.3 & $\downarrow$ 32.6 & $\downarrow$ 43.7 \\
        ~~+ SST-2 & $\downarrow$ 35.5 & $\downarrow$ 42.9 & $\downarrow$ 32.6 & $\downarrow$ 43.5 \\
        ~~+ MNLI & $\downarrow$ 34.8 & $\downarrow$ 41.8 & $\downarrow$ 32.4 & $\downarrow$ 43.3 \\
        \midrule
        \multicolumn{5}{c}{BART} \\
        \midrule
        rand. init & $\downarrow$ 25.3 & $\downarrow$ 12.5 & $\downarrow$ 23.2 & $\downarrow$ 12.2 \\
        PLM & $\bullet$ 39.1 & $\bullet$ 41.7 & $\bullet$ 31.8 & $\bullet$ 41.8 \\
        ~~+ CNN-DM & $\uparrow$ 40.9 & $\uparrow$ 44.3 & $\uparrow$ 32.7 & $\uparrow$ 42.8 \\
        ~~+ XSUM & $\uparrow$ 40.1 & $\uparrow$ 41.9 & $\uparrow$ 32.1 & $\downarrow$ 39.9 \\
        ~~+ SQuAD & $\uparrow$ 40.1 & $\uparrow$ 43.2 & $\downarrow$ 31.3 & $\downarrow$ 40.7 \\
        \midrule
        \multicolumn{5}{c}{Baselines} \\
        \midrule
        Right-Branch/Chain &  9.3 & 40.4 & 9.4 & 41.7 \\
        Left-Branch/Chain\textsuperscript{-1} & 7.5 & 12.7 & 1.5 & 12.2 \\
        Sum\textsubscript{CNN-DM}\shortcite{xiao2021predicting} & 21.4 & 20.5 & 17.6 & 15.8 \\
        Sum\textsubscript{NYT}\shortcite{xiao2021predicting} & 24.0 & 15.7 & 18.2 & 12.6 \\
        Two-Stage\textsubscript{RST-DT}\shortcite{wang2017two} & 72.0 & 71.2 & 54.0 & 54.5 \\
        Two-Stage\textsubscript{GUM} & 65.4 & 61.7 & 58.6 & 56.7 \\
        \bottomrule
    \end{tabular}}
    \caption{Original parseval (Span) and Unlabelled Attachment Score (UAS) of the single best performing self-attention matrix of the BERT and BART models compared with baselines and previous work. $\uparrow$, $\bullet$, $\downarrow$ indicate better, same, worse performance compared to the PLM. ``rand. init"=Randomly initialized transformer model of similar architecture as the PLM.}
    \label{tab:results}
\end{table}

\subsection{Discourse Quality}
We now focus on assessing the discourse information captured in the single best-performing self-attention head. In Table \ref{tab:results}, we compare the quality of generated discourse structures between different pre-trained and fine-tuned models, as well as additional baselines\footnote{For a more detailed analysis of the min., mean, median and max. self-attention performances see Appendix \ref{app:min_max}.}.
The results are separated into three sub-tables, showing the results for BERT, BART and baseline models on the RST-DT and GUM treebanks. In the BERT and BART sub-table, we further annotate each performance with $\uparrow$, $\bullet$, $\downarrow$, indicating the relative performance to the standard pre-trained model as superior, equal, or inferior.

Taking a look at the top sub-table (BERT) we find that, as expected, the randomly initialized transformer model achieves the worst performance. Fine-tuned models perform equal or worse than the standard PLM. Despite the inferior results of the fine-tuned models, the drop is rather small, with the sentiment analysis models consistently outperforming NLI. This seems reasonable, given that the sentiment analysis objective is intuitively more aligned with discourse structures (e.g., long-form reviews with potentially complex rhetorical structures) than the between-sentence NLI task, not involving multi-sentential text.

In the center sub-table (BART), a different trend emerges. While the worst performing model is still (as expected) the randomly initialized system, fine-tuned models mostly outperform the standard PLM. Interestingly, the model fine-tuned on the CNN-DM corpus consistently outperforms the BART baseline, while the XSUM model performs better on all but the GUM dependency structure evaluation. On one hand, the superior performance of both summarization models on the RST-DT dataset seems reasonable, given that the fine-tuning datasets and the evaluation treebank are both in the news domain. The strong results of the CNN-DM model on the GUM treebank, yet inferior performance of XSUM, potentially hints towards dependency discourse structures being less prominent when fine-tuning on the extreme summarization task, compared to the longer summaries in the CNN-DM corpus.
The question-answering task evaluated through the SQuAD fine-tuned model underperforms the standard PLM on GUM, however reaches superior performance on RST-DT. Since the SQuAD corpus is a subset of Wikipedia articles, more aligned with news articles than the 12 genres in GUM, we believe the stronger performance on RST-DT (i.e., news articles) is again reasonable, yet shows weaker generalization capabilities across domains (i.e., on the GUM corpus). Interestingly, the question-answering task seems more aligned with dependency than constituency trees, in line with what would be expected from a factoid-style question-answering model, focusing on important entities, rather than global constituency structures.

Directly comparing the BERT and BART models, the former performs better on three out of four metrics. At the same time, fine-tuning hurts the performance for BERT, however, improves BART models. Plausibly, these seemingly unintuitive results may be caused by the following co-occurring circumstances: (1) The inferior performance of BART can potentially be attributed to the decoder component capturing parts of the discourse structures, as well as the larger number of self-attention heads ``diluting'' the discourse information. (2) The different trends regarding fine-tuned models might be directly influenced by the input-length limitation to $512$ (BERT) and $1024$ (BART) sub-word tokens during the fine-tuning stage, hampering the ability to capture long-distance semantic and pragmatic relationships. This, in turn, limits the amount of discourse information captured, even for document-level datasets (e.g., Yelp, CNN-DM, SQuAD). With this restriction being more prominent in BERT, it potentially explains the comparably low performance of the fine-tuned models.

\begin{figure}[t]
    \centering
    \setlength{\belowcaptionskip}{-15pt}
    \includegraphics[width=\linewidth]{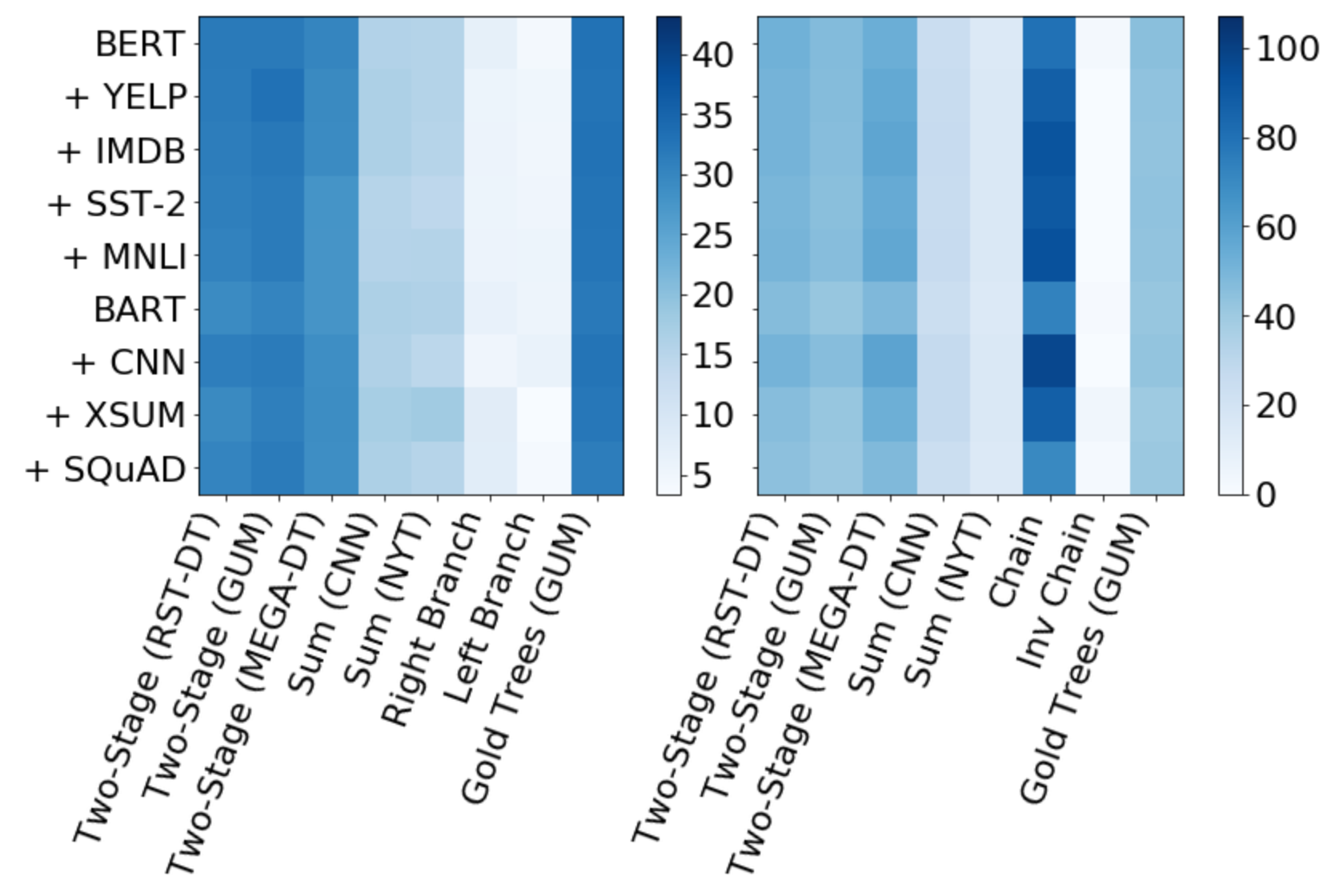}
     \caption{PLM discourse constituency (left) and dependency (right) structure overlap with baselines and gold trees (e.g., BERT $\leftrightarrow$ Two-Stage (RST-DT)) according to the original parseval and UAS metrics.}
    \label{fig:intersection_baselines}
\end{figure}

Finally, the bottom sub-table puts our results in the context of baselines. Compared to simple right- and left-branching trees (Span), the PLM-based models reach clearly superior performance. Looking at the chain/inverse chain structures (UAS), the improvements are generally lower, however, the vast majority still outperforms the baseline. Comparing the first two sub-tables against completely supervised methods (Two-Stage\textsubscript{RST-DT}, Two-Stage\textsubscript{GUM}), the BERT- and BART-based models are, unsurprisingly, inferior. Lastly, compared to the distantly supervised Sum\textsubscript{CNN-DM} and Sum\textsubscript{NYT} models, the PLM-based discourse performance shows clear improvements over the 6-layer, 8-head standard transformer.

\subsection{Discourse Similarity}
\label{sec:similarities}
Further exploring what kind of discourse information is captured in the PLM self-attention matrices, we directly compare the emergent discourse structures with baseline trees. This way, we aim to better understand if the information encapsulated in PLMs is complementary to existing methods, or if the PLMs only capture trivial discourse phenomena and simple biases (e.g., resemble right-branching constituency trees). Since the GUM dataset contains a more diverse set of test documents (12 genres) than the RST-DT corpus (news), we perform our experiments from here on exclusively on GUM.

\begin{figure}
    \centering
    \setlength{\belowcaptionskip}{-15pt}
    \subfigure[Head-aligned]{
    \includegraphics[width=.43\linewidth]{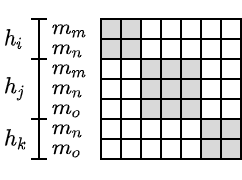}}
    \hfill
    \subfigure[Model-aligned]{
    \includegraphics[width=.43\linewidth]{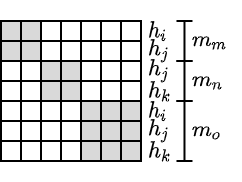}}
    \caption{Nested aggregation approach for discourse similarity. Grey cells contain same-head/same-model ((a)/(b)), white cells indicate between-head/between-model ((a)/(b)). Column indices equal row indices.}
    \label{fig:nested_aggregation}
\end{figure}

Figure \ref{fig:intersection_baselines} shows the micro-average structural overlap of discourse constituency (left) and dependency (right) trees between the PLM-generated structures and our baseline models, as well as gold-standard trees. Noticeably, the generated constituency trees (on the left) are most aligned with the structures predicted by supervised discourse parsers, showing only minimal overlap to simple structures (i.e., right- and left-branching trees). 
Taking a closer look at the generated dependency structures presented on the right side in Figure \ref{fig:intersection_baselines}, the alignment between PLM inferred discourse trees and the simple chain structure is predominant, suggesting a potential weakness in regards to the discourse captured in the BERT and BART model. Not surprisingly, the highest overlap between PLM-generated trees and the chain structure occurs when fine-tuning on the CNN-DM dataset, well-known to contain a strong lead-bias \cite{xing2021demoting}.

To better understand if the PLM-based discourse structures are complementary to existing, supervised discourse parsers, we further analyze the correctly predicted overlap. More specifically, we compute the intersection of both, the PLM model and the baselines with gold-standard trees (e.g., BERT $\cap$ Gold Trees $\leftrightarrow$ Two-Stage (RST-DT) $\cap$ Gold Trees) and further intersect the two resulting sets. This way, we explore if the correctly predicted PLM discourse structures are a subset of the correctly predicted trees by supervised approaches, or if complementary discourse information is captured. We find that >20\%/>16\% of the correctly predicted constituency/dependency structures of our PLM discourse inference approach are not captured by supervised models, making the exploration of ensemble methods a promising future avenue. A detailed version of Fig. \ref{fig:intersection_baselines} as well as more specific results regarding the correctly predicted overlap of discourse structures are shown in Appendix \ref{app:ensemble}. 

\begin{figure}
    \centering
    \setlength{\belowcaptionskip}{-15pt}
    \subfigure[Constituency Similarity]{
        \includegraphics[width=.48\linewidth]{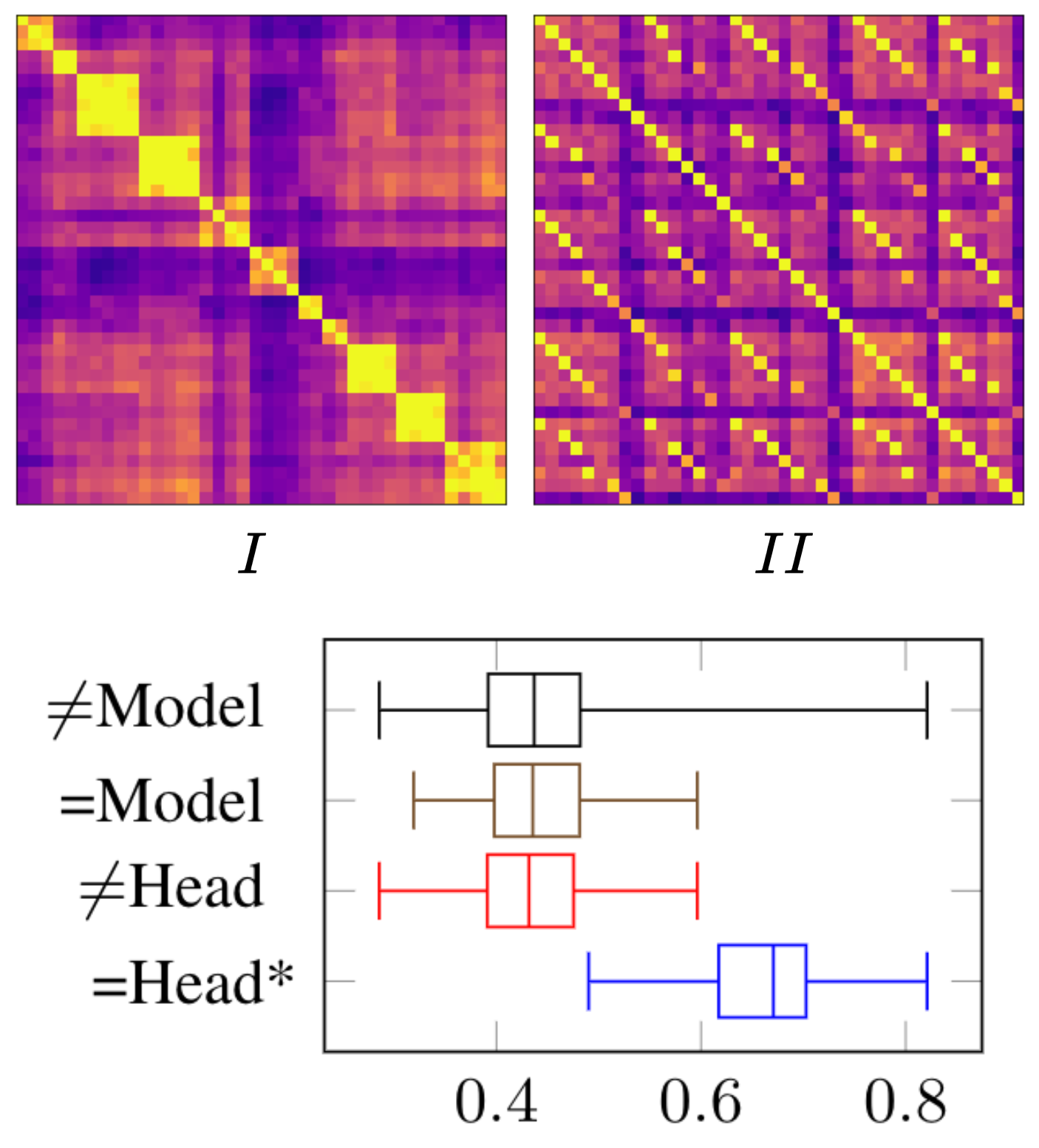}}
    \subfigure[Dependency Similarity]{
        \includegraphics[width=.48\linewidth]{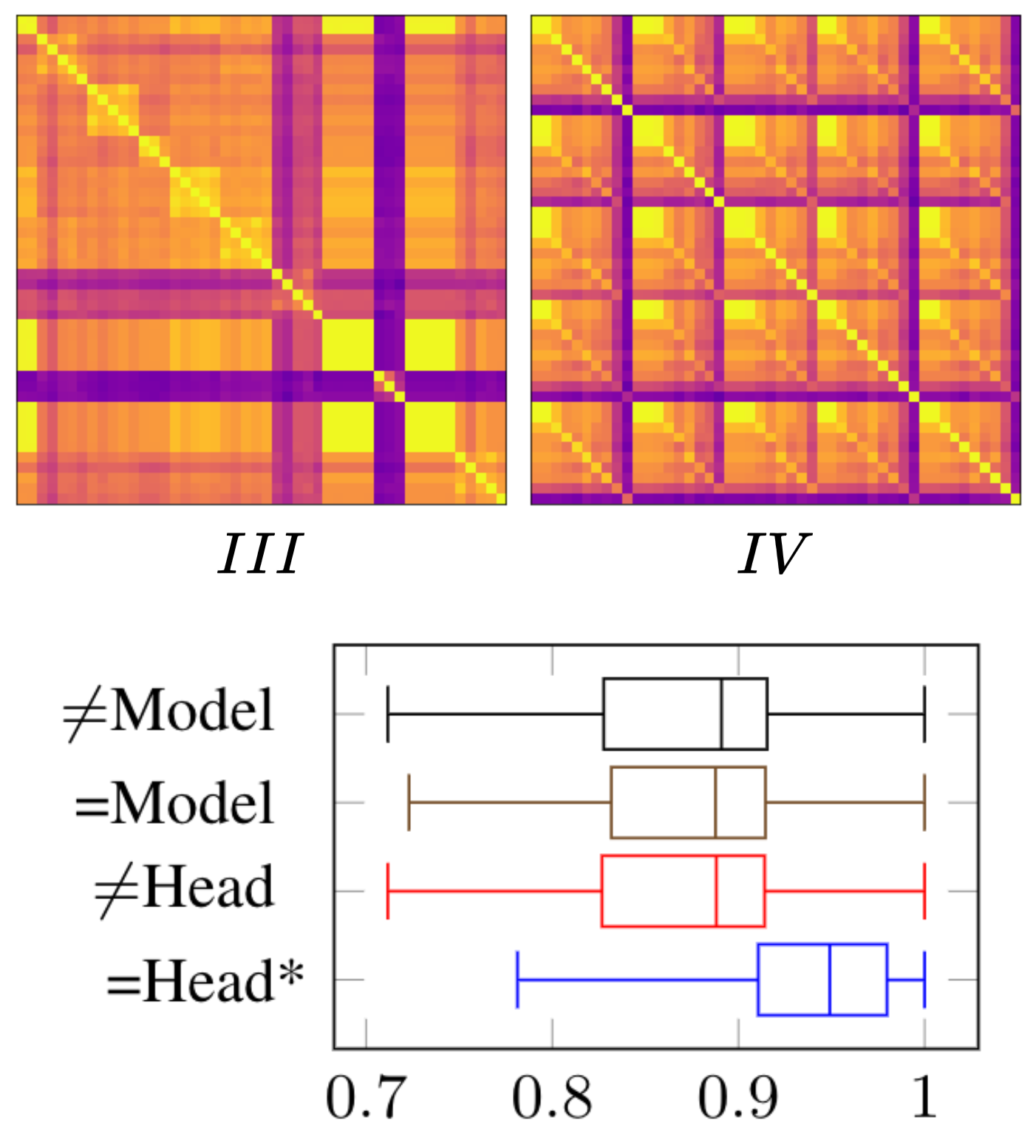}}
    \caption{BERT self-attention similarities on GUM.\\
    Top: Visual analysis of head-aligned ($I \& III$) and model-aligned ($II \& IV$) heatmaps. Yellow=high structural overlap, purple=low structural overlap.\\
    Bottom: Aggregated similarity of same heads, same models, different heads and different models showing the min, max and quartiles of the underlying distribution. *Significantly better than respective $\neq$Head/$\neq$Model performance with p-value $<0.05$.}
    \label{fig:bert_sim}
\end{figure}

\subsection{Discourse Redundancy}
\label{sec:head_layers}
Looking at the similarity of model self-attention heads in regards to their alignment with discourse information, we now explore if (1) the top performing heads $h_i, ..., h_k$ of a specific model $m_m$ capture redundant discourse structures, and if (2) the discourse information captured by a specific head $h_i$ across different models $m_m, ..., m_o$ contain similar discourse information.

Specifically, we pick the top $10$ best performing self-attention matrices of each model, remove self-attention heads that don't appear in at least two models (since no comparisons can be made), and compare the generated discourse structures in a nested aggregation approach.

Figure \ref{fig:nested_aggregation} shows a small-scale example of our nested visualization methodology. For the (self-attention) head-aligned approach (Figure \ref{fig:nested_aggregation}(a)), high similarity values along the diagonal (grey cells) would be expected if the same head $h_i$ encodes consistent discourse information across different fine-tuning tasks and datasets. Inversely, the model-aligned matrix (Figure \ref{fig:nested_aggregation}(b)) should show high values along the diagonal if different heads $h_i, ..., h_k$ in the same model $m_k$ capture redundant (i.e., similar) discourse information. Besides the visual inspection methodology presented in Figure \ref{fig:nested_aggregation}, we also compare aggregated similarities between the same head (=Head) against different heads ($\neq$Head) and between the same model (=Model) against different models ($\neq$Model)  (i.e., grey cells (=) and white cells ($\neq$) in Figure \ref{fig:nested_aggregation}(a) and (b)). In order to assess the statistical significance of the resulting differences in the underlying distributions, we compute a two-sided, independent t-test between same/different models and same/different heads\footnote{Prior to running the t-test we confirm similar variance and the assumption of normal distribution (Shapiro-Wilk test).}.

The resulting redundancy evaluations for BERT\footnote{Evaluations for BART can be found in Appendix \ref{app:self_attn_comp}.} are presented in Figure \ref{fig:bert_sim}. It appears that the same self-attention heads $h_i$ consistently encode similar discourse information across models indicated by: (1) High similarities (yellow) along the diagonal in heatmaps $I \& III$ and (2) through the statistically significant difference in distributions at the bottom of Figure \ref{fig:bert_sim}(a) and (b). However, different self-attention heads $h_i,...,h_k$ of the same model $m_m$ encode different discourse information (heatmaps $II \& IV$). While the trend is stronger for constituency tree structures, there is a single dependency self-attention head which does generally not align well between models and heads (purple line in heatmap $III$). Plausibly, this specific self-attention head encodes fine-tuning task specific discourse information. Overall, the similarity patterns observed in Figure \ref{fig:bert_sim}(a) and (b) point towards an opportunity to combine model self-attention heads to improve the discourse inference performance compared to the scores shown in Table \ref{tab:results}, where each self-attention head was assessed individually.

\section{Conclusions}
In this paper, we extend the line of \textit{BERTology} work by focusing on the important, yet less explored, alignment of pre-trained and fine-tuned PLMs with large-scale discourse structures. We propose a novel approach to infer discourse information for arbitrarily long documents. In our experiments, we find that the captured discourse information is local and general, even across a collection of fine-tuning tasks. We compare the inferred discourse trees with supervised, distantly supervised and simple baselines to explore the structural overlap, finding that constituency discourse trees align well with supervised models, however, contain complementary discourse information. Lastly, we individually explore self-attention matrices to analyze the information redundancy. We find that similar discourse information is consistently captured in the same heads. Based on the insights we gained in this analysis of large-scale discourse structures in PLMs, in the short term, we intend to explore new discourse inference methods using multiple (diverse) self-attention heads. Long term, we plan to analyze PLMs with enhanced input-length limitations. 

\section*{Acknowledgements}
We thank the anonymous reviewers and the UBC NLP group for their insightful comments and suggestions. This research was supported by the Language \& Speech Innovation Lab of Cloud BU, Huawei Technologies Co., Ltd and the Natural Sciences and Engineering Research Council of Canada (NSERC). Nous remercions le Conseil de recherches en sciences naturelles et en génie du Canada (CRSNG) de son soutien.

\bibliography{anthology,custom}

\begin{thebibliography}{60}
\expandafter\ifx\csname natexlab\endcsname\relax\def\natexlab#1{#1}\fi

\bibitem[{Adhikari et~al.(2019)Adhikari, Ram, Tang, and
  Lin}]{adhikari2019docbert}
Ashutosh Adhikari, Achyudh Ram, Raphael Tang, and Jimmy Lin. 2019.
\newblock Docbert: Bert for document classification.
\newblock \emph{arXiv preprint arXiv:1904.08398}.

\bibitem[{Beltagy et~al.(2020)Beltagy, Peters, and
  Cohan}]{beltagy2020longformer}
Iz~Beltagy, Matthew~E Peters, and Arman Cohan. 2020.
\newblock Longformer: The long-document transformer.
\newblock \emph{arXiv preprint arXiv:2004.05150}.

\bibitem[{Bhatia et~al.(2015)Bhatia, Ji, and Eisenstein}]{bhatia2015better}
Parminder Bhatia, Yangfeng Ji, and Jacob Eisenstein. 2015.
\newblock Better document-level sentiment analysis from rst discourse parsing.
\newblock In \emph{Proceedings of the 2015 Conference on Empirical Methods in
  Natural Language Processing}, pages 2212--2218.

\bibitem[{Carlson et~al.(2002)Carlson, Okurowski, and Marcu}]{carlson2002rst}
Lynn Carlson, Mary~Ellen Okurowski, and Daniel Marcu. 2002.
\newblock \emph{RST discourse treebank}.
\newblock Linguistic Data Consortium, University of Pennsylvania.

\bibitem[{Dai et~al.(2019)Dai, Yang, Yang, Carbonell, Le, and
  Salakhutdinov}]{dai2019transformer}
Zihang Dai, Zhilin Yang, Yiming Yang, Jaime~G Carbonell, Quoc Le, and Ruslan
  Salakhutdinov. 2019.
\newblock Transformer-xl: Attentive language models beyond a fixed-length
  context.
\newblock In \emph{Proceedings of the 57th Annual Meeting of the Association
  for Computational Linguistics}, pages 2978--2988.

\bibitem[{Devlin et~al.(2018)Devlin, Chang, Lee, and
  Toutanova}]{devlin2018bert}
Jacob Devlin, Ming-Wei Chang, Kenton Lee, and Kristina Toutanova. 2018.
\newblock Bert: Pre-training of deep bidirectional transformers for language
  understanding.
\newblock \emph{arXiv preprint arXiv:1810.04805}.

\bibitem[{Diao et~al.(2014)Diao, Qiu, Wu, Smola, Jiang, and
  Wang}]{diao2014jointly}
Qiming Diao, Minghui Qiu, Chao-Yuan Wu, Alexander~J Smola, Jing Jiang, and
  Chong Wang. 2014.
\newblock Jointly modeling aspects, ratings and sentiments for movie
  recommendation (jmars).
\newblock In \emph{Proceedings of the 20th ACM SIGKDD international conference
  on Knowledge discovery and data mining}, pages 193--202. ACM.

\bibitem[{Eisner(1996)}]{eisner-1996-three}
Jason~M. Eisner. 1996.
\newblock \href {https://aclanthology.org/C96-1058} {Three new probabilistic
  models for dependency parsing: An exploration}.
\newblock In \emph{{COLING} 1996 Volume 1: The 16th International Conference on
  Computational Linguistics}.

\bibitem[{Gerani et~al.(2019)Gerani, Carenini, and Ng}]{gerani2019modeling}
Shima Gerani, Giuseppe Carenini, and Raymond~T Ng. 2019.
\newblock Modeling content and structure for abstractive review summarization.
\newblock \emph{Computer Speech \& Language}, 53:302--331.

\bibitem[{Guz and Carenini(2020)}]{guz2020coreference}
Grigorii Guz and Giuseppe Carenini. 2020.
\newblock Coreference for discourse parsing: A neural approach.
\newblock In \emph{Proceedings of the First Workshop on Computational
  Approaches to Discourse}, pages 160--167.

\bibitem[{Guz et~al.(2020)Guz, Huber, and Carenini}]{guz2020unleashing}
Grigorii Guz, Patrick Huber, and Giuseppe Carenini. 2020.
\newblock Unleashing the power of neural discourse parsers-a context and
  structure aware approach using large scale pretraining.
\newblock In \emph{Proceedings of the 28th International Conference on
  Computational Linguistics}, pages 3794--3805.

\bibitem[{Hernault et~al.(2010)Hernault, Prendinger, Ishizuka
  et~al.}]{hernault2010hilda}
Hugo Hernault, Helmut Prendinger, Mitsuru Ishizuka, et~al. 2010.
\newblock Hilda: A discourse parser using support vector machine
  classification.
\newblock \emph{Dialogue \& Discourse}, 1(3).

\bibitem[{Hewitt and Manning(2019)}]{hewitt-manning-2019-structural}
John Hewitt and Christopher~D. Manning. 2019.
\newblock \href {https://doi.org/10.18653/v1/N19-1419} {{A} structural probe
  for finding syntax in word representations}.
\newblock In \emph{Proceedings of the 2019 Conference of the North {A}merican
  Chapter of the Association for Computational Linguistics: Human Language
  Technologies, Volume 1 (Long and Short Papers)}, pages 4129--4138,
  Minneapolis, Minnesota. Association for Computational Linguistics.

\bibitem[{Hogenboom et~al.(2015)Hogenboom, Frasincar, De~Jong, and
  Kaymak}]{hogenboom2015using}
Alexander Hogenboom, Flavius Frasincar, Franciska De~Jong, and Uzay Kaymak.
  2015.
\newblock Using rhetorical structure in sentiment analysis.
\newblock \emph{Commun. ACM}, 58(7):69--77.

\bibitem[{Huber and Carenini(2020)}]{huber2020mega}
Patrick Huber and Giuseppe Carenini. 2020.
\newblock Mega rst discourse treebanks with structure and nuclearity from
  scalable distant sentiment supervision.
\newblock In \emph{Proceedings of the 2020 Conference on Empirical Methods in
  Natural Language Processing (EMNLP)}, pages 7442--7457.

\bibitem[{Jawahar et~al.(2019)Jawahar, Sagot, and
  Seddah}]{jawahar-etal-2019-bert}
Ganesh Jawahar, Beno{\^\i}t Sagot, and Djam{\'e} Seddah. 2019.
\newblock \href {https://doi.org/10.18653/v1/P19-1356} {What does {BERT} learn
  about the structure of language?}
\newblock In \emph{Proceedings of the 57th Annual Meeting of the Association
  for Computational Linguistics}, pages 3651--3657, Florence, Italy.
  Association for Computational Linguistics.

\bibitem[{Ji and Eisenstein(2014)}]{ji2014representation}
Yangfeng Ji and Jacob Eisenstein. 2014.
\newblock Representation learning for text-level discourse parsing.
\newblock In \emph{Proceedings of the 52nd Annual Meeting of the Association
  for Computational Linguistics (Volume 1: Long Papers)}, volume~1, pages
  13--24.

\bibitem[{Ji and Smith(2017)}]{ji2017neural}
Yangfeng Ji and Noah~A Smith. 2017.
\newblock Neural discourse structure for text categorization.
\newblock In \emph{Proceedings of the 55th Annual Meeting of the Association
  for Computational Linguistics (Volume 1: Long Papers)}, pages 996--1005.

\bibitem[{Joshi et~al.(2020)Joshi, Chen, Liu, Weld, Zettlemoyer, and
  Levy}]{joshi2020spanbert}
Mandar Joshi, Danqi Chen, Yinhan Liu, Daniel~S Weld, Luke Zettlemoyer, and Omer
  Levy. 2020.
\newblock Spanbert: Improving pre-training by representing and predicting
  spans.
\newblock \emph{Transactions of the Association for Computational Linguistics},
  8:64--77.

\bibitem[{Joty et~al.(2015)Joty, Carenini, and Ng}]{joty2015codra}
Shafiq Joty, Giuseppe Carenini, and Raymond~T Ng. 2015.
\newblock Codra: A novel discriminative framework for rhetorical analysis.
\newblock \emph{Computational Linguistics}, 41(3).

\bibitem[{Jurafsky and Martin(2014)}]{jurafsky2014speech}
Dan Jurafsky and James~H Martin. 2014.
\newblock \emph{Speech and language processing}, volume~3.
\newblock Pearson London.

\bibitem[{Kim et~al.(2019)Kim, Choi, Edmiston, and Lee}]{kim2019pre}
Taeuk Kim, Jihun Choi, Daniel Edmiston, and Sang-goo Lee. 2019.
\newblock Are pre-trained language models aware of phrases? simple but strong
  baselines for grammar induction.
\newblock In \emph{International Conference on Learning Representations}.

\bibitem[{Kitaev et~al.(2020)Kitaev, Kaiser, and Levskaya}]{kitaev2020reformer}
Nikita Kitaev, {\L}ukasz Kaiser, and Anselm Levskaya. 2020.
\newblock Reformer: The efficient transformer.
\newblock \emph{arXiv preprint arXiv:2001.04451}.

\bibitem[{Kobayashi et~al.(2020)Kobayashi, Hirao, Kamigaito, Okumura, and
  Nagata}]{kobayashi2020top}
Naoki Kobayashi, Tsutomu Hirao, Hidetaka Kamigaito, Manabu Okumura, and Masaaki
  Nagata. 2020.
\newblock Top-down rst parsing utilizing granularity levels in documents.
\newblock In \emph{Proceedings of the AAAI Conference on Artificial
  Intelligence}, volume~34, pages 8099--8106.

\bibitem[{Kobayashi et~al.(2019)Kobayashi, Hirao, Nakamura, Kamigaito, Okumura,
  and Nagata}]{kobayashi2019split}
Naoki Kobayashi, Tsutomu Hirao, Kengo Nakamura, Hidetaka Kamigaito, Manabu
  Okumura, and Masaaki Nagata. 2019.
\newblock Split or merge: Which is better for unsupervised rst parsing?
\newblock In \emph{Proceedings of the 2019 Conference on Empirical Methods in
  Natural Language Processing and the 9th International Joint Conference on
  Natural Language Processing (EMNLP-IJCNLP)}, pages 5797--5802.

\bibitem[{Koto et~al.(2021{\natexlab{a}})Koto, Lau, and
  Baldwin}]{koto2021discourse}
Fajri Koto, Jey~Han Lau, and Timothy Baldwin. 2021{\natexlab{a}}.
\newblock Discourse probing of pretrained language models.
\newblock In \emph{Proceedings of the 2021 Conference of the North American
  Chapter of the Association for Computational Linguistics: Human Language
  Technologies}, pages 3849--3864.

\bibitem[{Koto et~al.(2021{\natexlab{b}})Koto, Lau, and Baldwin}]{koto2021top}
Fajri Koto, Jey~Han Lau, and Timothy Baldwin. 2021{\natexlab{b}}.
\newblock Top-down discourse parsing via sequence labelling.
\newblock In \emph{Proceedings of the 16th Conference of the European Chapter
  of the Association for Computational Linguistics: Main Volume}, pages
  715--726.

\bibitem[{Kurfal{\i} and {\"O}stling(2021)}]{kurfali2021probing}
Murathan Kurfal{\i} and Robert {\"O}stling. 2021.
\newblock Probing multilingual language models for discourse.
\newblock \emph{arXiv preprint arXiv:2106.04832}.

\bibitem[{Lewis et~al.(2020)Lewis, Liu, Goyal, Ghazvininejad, Mohamed, Levy,
  Stoyanov, and Zettlemoyer}]{lewis2020bart}
Mike Lewis, Yinhan Liu, Naman Goyal, Marjan Ghazvininejad, Abdelrahman Mohamed,
  Omer Levy, Veselin Stoyanov, and Luke Zettlemoyer. 2020.
\newblock Bart: Denoising sequence-to-sequence pre-training for natural
  language generation, translation, and comprehension.
\newblock In \emph{Proceedings of the 58th Annual Meeting of the Association
  for Computational Linguistics}, pages 7871--7880.

\bibitem[{Li et~al.(2014)Li, Wang, Cao, and Li}]{li-etal-2014-text}
Sujian Li, Liang Wang, Ziqiang Cao, and Wenjie Li. 2014.
\newblock \href {https://doi.org/10.3115/v1/P14-1003} {Text-level discourse
  dependency parsing}.
\newblock In \emph{Proceedings of the 52nd Annual Meeting of the Association
  for Computational Linguistics (Volume 1: Long Papers)}, pages 25--35,
  Baltimore, Maryland. Association for Computational Linguistics.

\bibitem[{Liu et~al.(2019)Liu, Ott, Goyal, Du, Joshi, Chen, Levy, Lewis,
  Zettlemoyer, and Stoyanov}]{liu2019roberta}
Yinhan Liu, Myle Ott, Naman Goyal, Jingfei Du, Mandar Joshi, Danqi Chen, Omer
  Levy, Mike Lewis, Luke Zettlemoyer, and Veselin Stoyanov. 2019.
\newblock Roberta: A robustly optimized bert pretraining approach.
\newblock \emph{arXiv preprint arXiv:1907.11692}.

\bibitem[{Mann and Thompson(1988)}]{mann1988rhetorical}
William~C Mann and Sandra~A Thompson. 1988.
\newblock Rhetorical structure theory: Toward a functional theory of text
  organization.
\newblock \emph{Text-Interdisciplinary Journal for the Study of Discourse},
  8(3):243--281.

\bibitem[{Mare{\v{c}}ek and Rosa(2019)}]{marevcek2019balustrades}
David Mare{\v{c}}ek and Rudolf Rosa. 2019.
\newblock From balustrades to pierre vinken: Looking for syntax in transformer
  self-attentions.
\newblock In \emph{Proceedings of the 2019 ACL Workshop BlackboxNLP: Analyzing
  and Interpreting Neural Networks for NLP}, pages 263--275.

\bibitem[{Michael et~al.(2020)Michael, Botha, and
  Tenney}]{michael-etal-2020-asking}
Julian Michael, Jan~A. Botha, and Ian Tenney. 2020.
\newblock \href {https://doi.org/10.18653/v1/2020.emnlp-main.552} {Asking
  without telling: Exploring latent ontologies in contextual representations}.
\newblock In \emph{Proceedings of the 2020 Conference on Empirical Methods in
  Natural Language Processing (EMNLP)}, pages 6792--6812, Online. Association
  for Computational Linguistics.

\bibitem[{Morey et~al.(2017)Morey, Muller, and Asher}]{morey2017much}
Mathieu Morey, Philippe Muller, and Nicholas Asher. 2017.
\newblock How much progress have we made on rst discourse parsing? a
  replication study of recent results on the rst-dt.
\newblock In \emph{Proceedings of the 2017 Conference on Empirical Methods in
  Natural Language Processing}, pages 1319--1324.

\bibitem[{Nallapati et~al.(2016)Nallapati, Zhou, dos Santos,
  GuÌ‡l{\c{c}}ehre, and Xiang}]{nallapati-etal-2016-abstractive}
Ramesh Nallapati, Bowen Zhou, Cicero dos Santos, {\c{C}}a{\u{g}}lar
  GuÌ‡l{\c{c}}ehre, and Bing Xiang. 2016.
\newblock \href {https://doi.org/10.18653/v1/K16-1028} {Abstractive text
  summarization using sequence-to-sequence {RNN}s and beyond}.
\newblock In \emph{Proceedings of The 20th {SIGNLL} Conference on Computational
  Natural Language Learning}, pages 280--290, Berlin, Germany. Association for
  Computational Linguistics.

\bibitem[{Narayan et~al.(2018)Narayan, Cohen, and Lapata}]{Narayan2018DontGM}
Shashi Narayan, Shay~B. Cohen, and Mirella Lapata. 2018.
\newblock Don't give me the details, just the summary! topic-aware
  convolutional neural networks for extreme summarization.
\newblock \emph{ArXiv}, abs/1808.08745.

\bibitem[{Nejat et~al.(2017)Nejat, Carenini, and Ng}]{nejat2017exploring}
Bita Nejat, Giuseppe Carenini, and Raymond Ng. 2017.
\newblock Exploring joint neural model for sentence level discourse parsing and
  sentiment analysis.
\newblock In \emph{Proceedings of the 18th Annual SIGdial Meeting on Discourse
  and Dialogue}, pages 289--298.

\bibitem[{Nguyen et~al.(2021)Nguyen, Nguyen, Joty, and Li}]{nguyen2021rst}
Thanh-Tung Nguyen, Xuan-Phi Nguyen, Shafiq Joty, and Xiaoli Li. 2021.
\newblock Rst parsing from scratch.
\newblock In \emph{Proceedings of the 2021 Conference of the North American
  Chapter of the Association for Computational Linguistics: Human Language
  Technologies}, pages 1613--1625.

\bibitem[{O{\u{g}}uz et~al.(2021)O{\u{g}}uz, Lakhotia, Gupta, Lewis, Karpukhin,
  Piktus, Chen, Riedel, Yih, Gupta et~al.}]{ouguz2021domain}
Barlas O{\u{g}}uz, Kushal Lakhotia, Anchit Gupta, Patrick Lewis, Vladimir
  Karpukhin, Aleksandra Piktus, Xilun Chen, Sebastian Riedel, Wen-tau Yih,
  Sonal Gupta, et~al. 2021.
\newblock Domain-matched pre-training tasks for dense retrieval.
\newblock \emph{arXiv preprint arXiv:2107.13602}.

\bibitem[{Pandia et~al.(2021)Pandia, Cong, and Ettinger}]{pandia2021pragmatic}
Lalchand Pandia, Yan Cong, and Allyson Ettinger. 2021.
\newblock Pragmatic competence of pre-trained language models through the lens
  of discourse connectives.
\newblock \emph{arXiv preprint arXiv:2109.12951}.

\bibitem[{Papanikolaou et~al.(2019)Papanikolaou, Roberts, and
  Pierleoni}]{papanikolaou-etal-2019-deep}
Yannis Papanikolaou, Ian Roberts, and Andrea Pierleoni. 2019.
\newblock \href {https://doi.org/10.18653/v1/D19-6108} {Deep bidirectional
  transformers for relation extraction without supervision}.
\newblock In \emph{Proceedings of the 2nd Workshop on Deep Learning Approaches
  for Low-Resource NLP (DeepLo 2019)}, pages 67--75, Hong Kong, China.
  Association for Computational Linguistics.

\bibitem[{Prasad et~al.(2008)Prasad, Dinesh, Lee, Miltsakaki, Robaldo, Joshi,
  and Webber}]{prasadpenn}
Rashmi Prasad, Nikhil Dinesh, Alan Lee, Eleni Miltsakaki, Livio Robaldo,
  Aravind Joshi, and Bonnie Webber. 2008.
\newblock The penn discourse treebank 2.0.
\newblock \emph{LREC}.

\bibitem[{Rae et~al.(2019)Rae, Potapenko, Jayakumar, and
  Lillicrap}]{rae2019compressive}
Jack~W Rae, Anna Potapenko, Siddhant~M Jayakumar, and Timothy~P Lillicrap.
  2019.
\newblock Compressive transformers for long-range sequence modelling.
\newblock \emph{arXiv preprint arXiv:1911.05507}.

\bibitem[{Raganato and Tiedemann(2018)}]{raganato2018analysis}
Alessandro Raganato and J{\"o}rg Tiedemann. 2018.
\newblock An analysis of encoder representations in transformer-based machine
  translation.
\newblock In \emph{Proceedings of the 2018 EMNLP Workshop BlackboxNLP:
  Analyzing and Interpreting Neural Networks for NLP}, pages 287--297.

\bibitem[{Rajpurkar et~al.(2016)Rajpurkar, Zhang, Lopyrev, and
  Liang}]{rajpurkar2016squad}
Pranav Rajpurkar, Jian Zhang, Konstantin Lopyrev, and Percy Liang. 2016.
\newblock Squad: 100,000+ questions for machine comprehension of text.
\newblock In \emph{Proceedings of the 2016 Conference on Empirical Methods in
  Natural Language Processing}, pages 2383--2392.

\bibitem[{Rogers et~al.(2020)Rogers, Kovaleva, and
  Rumshisky}]{rogers2020primer}
Anna Rogers, Olga Kovaleva, and Anna Rumshisky. 2020.
\newblock A primer in bertology: What we know about how bert works.
\newblock \emph{Transactions of the Association for Computational Linguistics},
  8:842--866.

\bibitem[{Socher et~al.(2013)Socher, Perelygin, Wu, Chuang, Manning, Ng, and
  Potts}]{socher-etal-2013-recursive}
Richard Socher, Alex Perelygin, Jean Wu, Jason Chuang, Christopher~D. Manning,
  Andrew Ng, and Christopher Potts. 2013.
\newblock \href {https://www.aclweb.org/anthology/D13-1170} {Recursive deep
  models for semantic compositionality over a sentiment treebank}.
\newblock In \emph{Proceedings of the 2013 Conference on Empirical Methods in
  Natural Language Processing}, pages 1631--1642, Seattle, Washington, USA.
  Association for Computational Linguistics.

\bibitem[{Vaswani et~al.(2017)Vaswani, Shazeer, Parmar, Uszkoreit, Jones,
  Gomez, Kaiser, and Polosukhin}]{vaswani2017attention}
Ashish Vaswani, Noam Shazeer, Niki Parmar, Jakob Uszkoreit, Llion Jones,
  Aidan~N Gomez, {\L}ukasz Kaiser, and Illia Polosukhin. 2017.
\newblock Attention is all you need.
\newblock In \emph{Advances in neural information processing systems}, pages
  5998--6008.

\bibitem[{Wang et~al.(2017)Wang, Li, and Wang}]{wang2017two}
Yizhong Wang, Sujian Li, and Houfeng Wang. 2017.
\newblock A two-stage parsing method for text-level discourse analysis.
\newblock In \emph{Proceedings of the 55th Annual Meeting of the Association
  for Computational Linguistics (Volume 2: Short Papers)}, pages 184--188.

\bibitem[{Williams et~al.(2018)Williams, Nangia, and Bowman}]{N18-1101}
Adina Williams, Nikita Nangia, and Samuel Bowman. 2018.
\newblock \href {http://aclweb.org/anthology/N18-1101} {A broad-coverage
  challenge corpus for sentence understanding through inference}.
\newblock In \emph{Proceedings of the 2018 Conference of the North American
  Chapter of the Association for Computational Linguistics: Human Language
  Technologies, Volume 1 (Long Papers)}, pages 1112--1122. Association for
  Computational Linguistics.

\bibitem[{Wu et~al.(2020)Wu, Chen, Kao, and Liu}]{wu2020perturbed}
Zhiyong Wu, Yun Chen, Ben Kao, and Qun Liu. 2020.
\newblock Perturbed masking: Parameter-free probing for analyzing and
  interpreting bert.
\newblock In \emph{Proceedings of the 58th Annual Meeting of the Association
  for Computational Linguistics}, pages 4166--4176.

\bibitem[{Xiao et~al.(2021{\natexlab{a}})Xiao, Beltagy, Carenini, and
  Cohan}]{xiao2021primer}
Wen Xiao, Iz~Beltagy, Giuseppe Carenini, and Arman Cohan. 2021{\natexlab{a}}.
\newblock Primer: Pyramid-based masked sentence pre-training for multi-document
  summarization.
\newblock \emph{arXiv preprint arXiv:2110.08499}.

\bibitem[{Xiao et~al.(2021{\natexlab{b}})Xiao, Huber, and
  Carenini}]{xiao2021predicting}
Wen Xiao, Patrick Huber, and Giuseppe Carenini. 2021{\natexlab{b}}.
\newblock Predicting discourse trees from transformer-based neural summarizers.
\newblock In \emph{Proceedings of the 2021 Conference of the North American
  Chapter of the Association for Computational Linguistics: Human Language
  Technologies}, pages 4139--4152.

\bibitem[{Xing et~al.(2021)Xing, Xiao, and Carenini}]{xing2021demoting}
Linzi Xing, Wen Xiao, and Giuseppe Carenini. 2021.
\newblock Demoting the lead bias in news summarization via alternating
  adversarial learning.
\newblock \emph{arXiv preprint arXiv:2105.14241}.

\bibitem[{Yang et~al.(2019)Yang, Dai, Yang, Carbonell, Salakhutdinov, and
  Le}]{yang2019xlnet}
Zhilin Yang, Zihang Dai, Yiming Yang, Jaime Carbonell, Russ~R Salakhutdinov,
  and Quoc~V Le. 2019.
\newblock Xlnet: Generalized autoregressive pretraining for language
  understanding.
\newblock \emph{Advances in neural information processing systems}, 32.

\bibitem[{Zeldes(2017)}]{Zeldes2017}
Amir Zeldes. 2017.
\newblock \href {https://doi.org/http://dx.doi.org/10.1007/s10579-016-9343-x}
  {The {GUM} corpus: Creating multilayer resources in the classroom}.
\newblock \emph{Language Resources and Evaluation}, 51(3):581--612.

\bibitem[{Zhang et~al.(2020)Zhang, Zhao, Saleh, and Liu}]{zhang2020pegasus}
Jingqing Zhang, Yao Zhao, Mohammad Saleh, and Peter Liu. 2020.
\newblock Pegasus: Pre-training with extracted gap-sentences for abstractive
  summarization.
\newblock In \emph{International Conference on Machine Learning}, pages
  11328--11339. PMLR.

\bibitem[{Zhang et~al.(2015)Zhang, Zhao, and LeCun}]{zhang2015character}
Xiang Zhang, Junbo Zhao, and Yann LeCun. 2015.
\newblock Character-level convolutional networks for text classification.
\newblock \emph{arXiv preprint arXiv:1502.01710}.

\bibitem[{Zhu et~al.(2020)Zhu, Pan, Abdalla, and Rudzicz}]{zhu2020examining}
Zining Zhu, Chuer Pan, Mohamed Abdalla, and Frank Rudzicz. 2020.
\newblock Examining the rhetorical capacities of neural language models.
\newblock In \emph{Proceedings of the Third BlackboxNLP Workshop on Analyzing
  and Interpreting Neural Networks for NLP}, pages 16--32.

\end{thebibliography}
\bibliographystyle{acl_natbib}

\appendix
\onecolumn

\section{Huggingface Models}
\label{app:links}
We investigate 7 fine-tuned BERT and BART models from the huggingface model library, as well as the two pre-trained models. The model names and links are provided in Table \ref{tab:links}
\begin{table*}[h]
    \centering
    \resizebox{\linewidth}{!}{
    \begin{tabular}{lll}
        \toprule
        Pre-Trained & Fine-Tuned & Link \\
        \midrule
        BERT-base &  -- &  \url{https://huggingface.co/bert-base-uncased}\\
        BERT-base &  IMDB & \url{https://huggingface.co/textattack/bert-base-uncased-imdb} \\
        BERT-base &  Yelp & \url{https://huggingface.co/fabriceyhc/bert-base-uncased-yelp_polarity} \\
        BERT-base &  SST-2 & \url{https://huggingface.co/textattack/bert-base-uncased-SST-2} \\
        BERT-base &  MNLI & \url{https://huggingface.co/textattack/bert-base-uncased-MNLI} \\
        \midrule
        BART-large &  -- & \url{https://huggingface.co/facebook/bart-large} \\
        BART-large &  CNN-DM & \url{https://huggingface.co/facebook/bart-large-cnn} \\
        BART-large &  XSUM & \url{https://huggingface.co/facebook/bart-large-xsum} \\
        BART-large &  SQuAD & \url{https://huggingface.co/valhalla/bart-large-finetuned-squadv1} \\
        \bottomrule
    \end{tabular}}
    \caption{Huggingface pre-trained and fine-tuned model links.}
    \label{tab:links}
\end{table*}

\section{Test-Set Results on RST-DT and GUM}
\label{sec:app_self_attn_test}

\begin{figure}[h]
    \centering
    \subfigure[BERT: PLM, +IMDB, +Yelp, +MNLI, +SST-2]{
        \includegraphics[width=.75\linewidth]{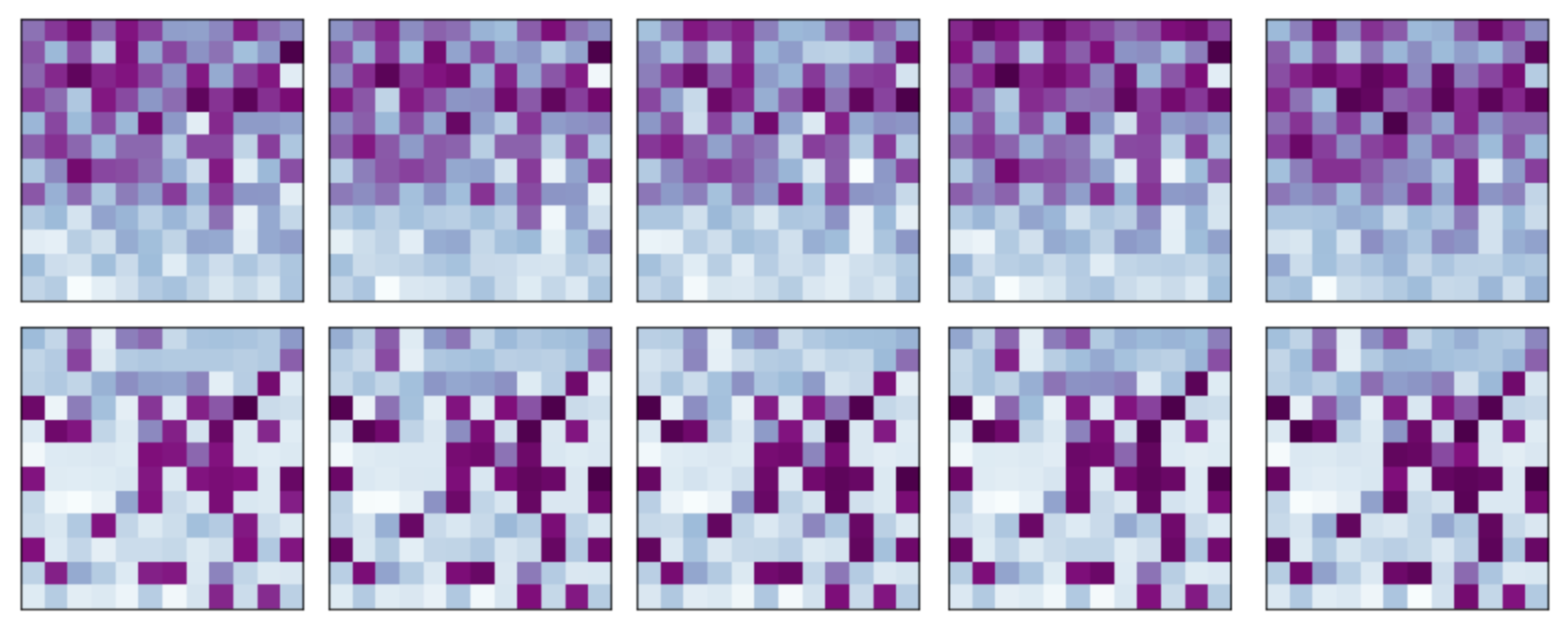}}
    \centering
    \subfigure[BART: PLM, +CNN-DM, +XSUM, +SQuAD]{
    \includegraphics[width=.75\linewidth]{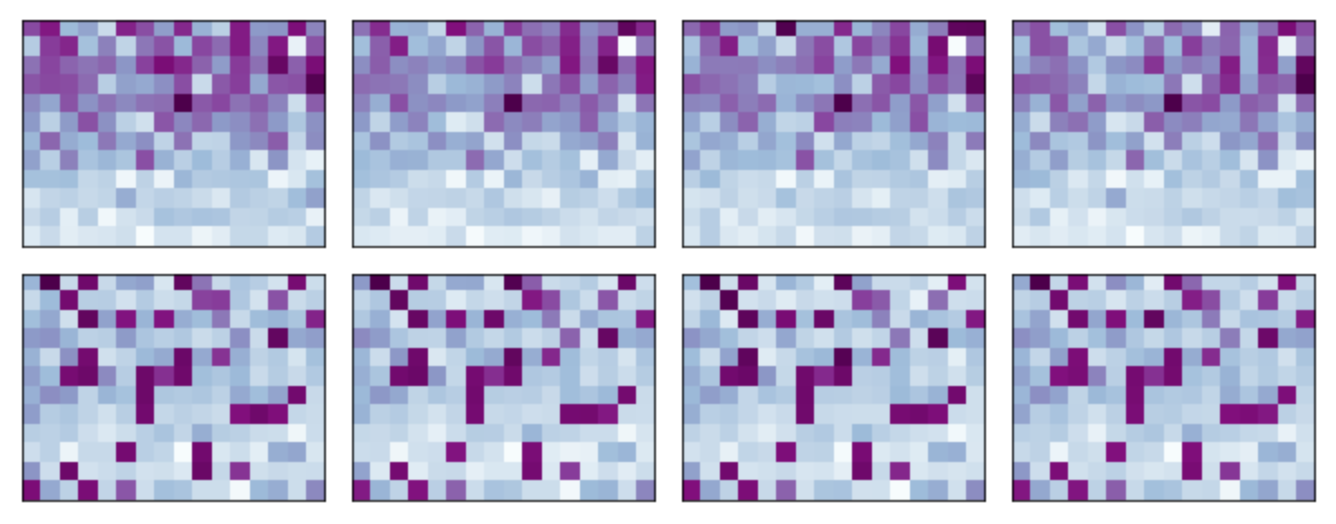}}
    \caption{Constituency (top) and dependency (bottom) discourse tree evaluation of BERT (a) and BART (b) models on RST-DT (test). Purple=high score, blue=low score. + indicates fine-tuning dataset.}
    \label{fig:rst_full_heatmap_collection}
\end{figure}

\begin{figure}[ht]
    \centering
    \subfigure[BERT: PLM, +IMDB, +Yelp, +MNLI, +SST-2]{
        \includegraphics[width=.75\linewidth]{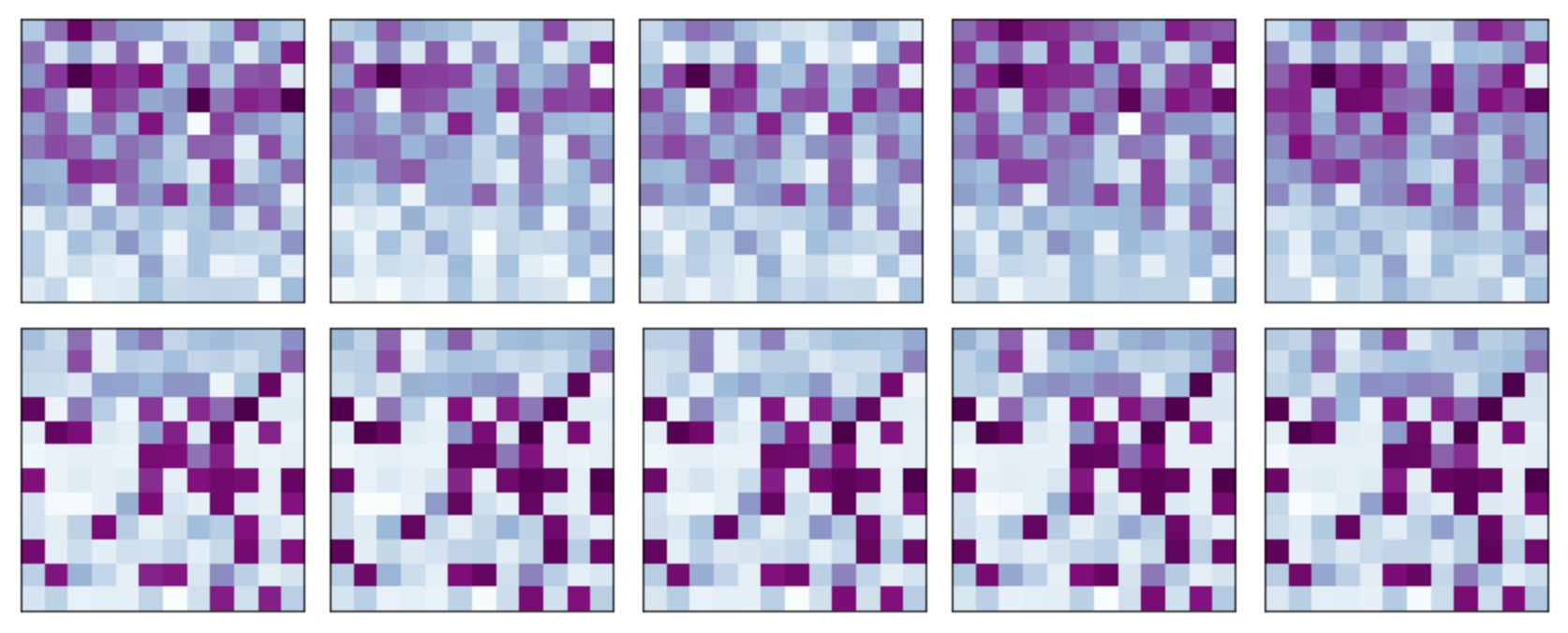}}
    \centering
    \subfigure[BART: PLM, +CNN-DM, +XSUM, +SQuAD]{
    \includegraphics[width=.75\linewidth]{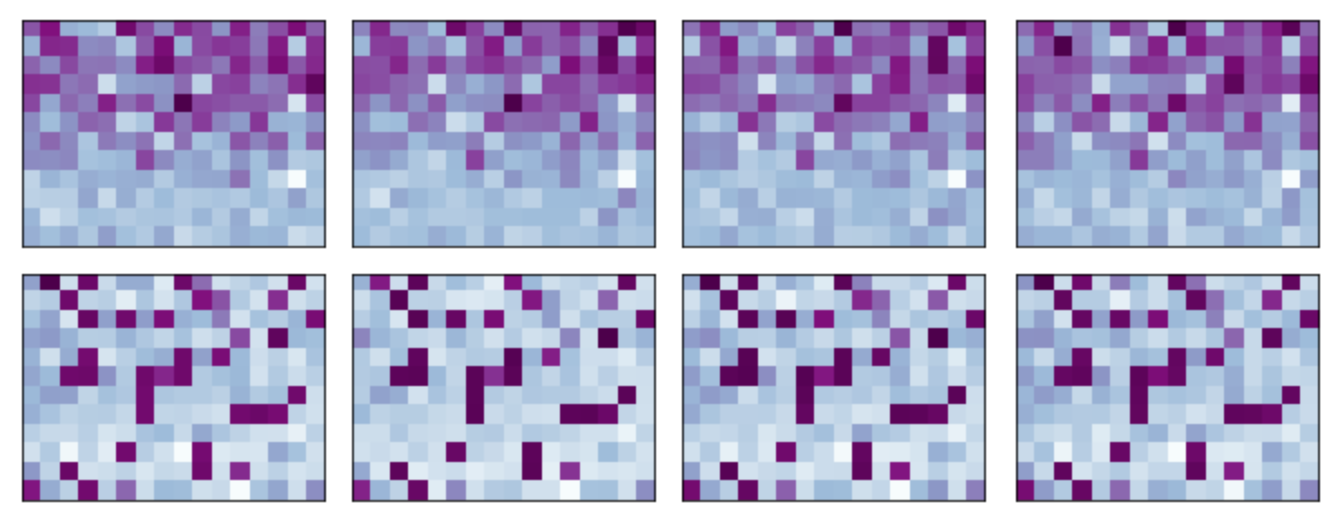}}
    \caption{Constituency (top) and dependency (bottom) discourse tree evaluation of BERT (a) and BART (b) models on GUM (test). Purple=high score, blue=low score. + indicates fine-tuning dataset.}
    \label{fig:gum_full_heatmap_collection}
\end{figure}

\clearpage
\newpage
\section{Detailed Self-Attention Statistics}
\label{app:min_max}

\begin{table}[h]
    \centering
    \scalebox{.9}{
    \begin{tabular}{lrrrrrrrrr}
        \toprule
        \multirow{2}{*}{Model} & \multicolumn{4}{c}{Span} & & \multicolumn{4}{c}{Eisner} \\
         & Min & Med & Mean & Max & & Min & Med & Mean & Max \\
         \midrule
         \multicolumn{10}{c}{RST-DT} \\
         \midrule
         rand. init & 21.7 & 23.4 & 23.4 & 25.5 & & 7.5 & 10.3 & 10.3 & 13.3 \\
         PLM & 19.3 & 27.0 & 27.4 & 35.7 & & 6.6 & 17.4 & 21.6 & 45.3 \\
         ~~+ IMDB & 19.7 & 26.9 & 27.2 & 35.4 & & 6.6 & 16.9 & 21.3 & 42.8 \\
         ~~+ YELP & 20.2 & 26.6 & 26.9 & 34.7 & & 7.0 & 16.5 & 21.0 & 42.3 \\
         ~~+ SST-2 & 19.5 & 27.3 & 27.7 & 35.5 & & 7.3 & 17.6 & 21.9 & 42.9 \\
         ~~+ MNLI & 18.5 & 26.9 & 27.1 & 34.8 & & 6.9 & 17.5 & 21.5 & 41.8 \\
         \midrule
         \multicolumn{10}{c}{GUM} \\
         \midrule
         rand. init & 18.6 & 21.0 & 21.0 & 23.2 & & 7.9 & 10.1 & 10.1 & 12.4 \\
         PLM & 17.8 & 24.2 & 24.3 & 32.6 & & 6.7 & 16.0 & 21.2 & 45.2 \\
         ~~+ IMDB & 18.1 & 23.8 & 24.1 & 32.7 & & 6.1 & 15.9 & 21.0 & 43.3 \\
         ~~+ YELP & 18.6 & 24.0 & 23.9 & 32.3 & & 7.0 & 15.8 & 20.7 & 43.7 \\
         ~~+ SST-2 & 18.2 & 24.6 & 24.7 & 32.3 & & 6.5 & 16.5 & 21.6 & 43.5 \\
         ~~+ MNLI & 17.4 & 23.9 & 24.2 & 32.1 & & 6.8 & 16.6 & 21.3 & 43.3 \\
         \bottomrule
    \end{tabular}}
    \caption{Minimum, median, mean and maximum performance of the self-attention matrices on RST-DT and GUM for the BERT model.}
    \label{tab:min_max_BERT}
\end{table}

\begin{table}[h]
    \centering
    \scalebox{.9}{
    \begin{tabular}{lrrrrrrrrr}
        \toprule
        \multirow{2}{*}{Model} & \multicolumn{4}{c}{Span} & & \multicolumn{4}{c}{Eisner} \\
         & Min & Med & Mean & Max & & Min & Med & Mean & Max \\
         \midrule
         \multicolumn{10}{c}{RST-DT} \\
         \midrule
         rand. init & 20.3 & 23.3 & 23.3 & 25.3 & & 8.5 & 10.6 & 10.6 & 12.5 \\
         PLM & 20.3 & 28.3 & 28.5 & 39.1 & & 4.1 & 15.8 & 19.2 & 41.7 \\
         ~~+ CNN-DM & 20.5 & 28.6 & 28.7 & 40.9 & & 3.6 & 15.2 & 19.2 & 44.3 \\
         ~~+ XSUM & 20.2 & 27.6 & 28.3 & 40.1 & & 4.8 & 14.8 & 18.7 & 41.9 \\
         ~~+ SQuAD & 20.5 & 27.6 & 28.2 & 40.1 & & 2.8 & 14.8 & 18.8 & 43.2 \\
         \midrule
         \multicolumn{10}{c}{GUM} \\
         \midrule
         rand. init & 18.6 & 21.0 & 21.0 & 23.2 & & 8.0 & 10.2 & 10.2 & 12.2 \\
         PLM & 16.7 & 23.4 & 23.8 & 31.5 & & 2.6 & 15.2 & 18.7 & 41.8 \\
         ~~+ CNN-DM & 15.9 & 23.7 & 24.1 & 32.4 & & 3.7 & 14.7 & 18.9 & 42.8 \\
         ~~+ XSUM & 16.4 & 23.2 & 23.9 & 31.8 & & 3.0 & 14.1 & 18.1 & 39.9 \\
         ~~+ SQuAD & 16.1 & 23.4 & 23.8 & 31.0 & & 2.4 & 14.8 & 18.3 & 40.7 \\
         \bottomrule
    \end{tabular}}
    \caption{Minimum, median, mean and maximum performance of the self-attention matrices on RST-DT and GUM for the BART model.}
    \label{tab:min_max_BART}
\end{table}

\newpage
\section{Details of Structural Discourse Similarity}
\label{app:ensemble}
\begin{figure*}[h]
    \centering
    \includegraphics[width=.46\textwidth]{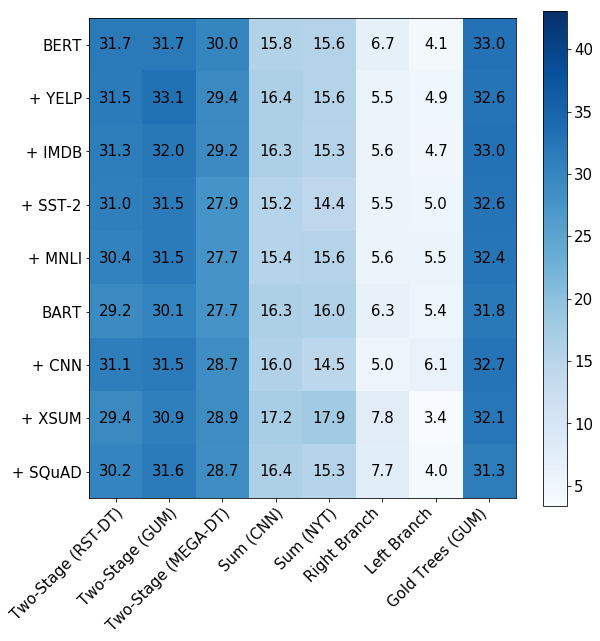}
    \includegraphics[width=.46\textwidth]{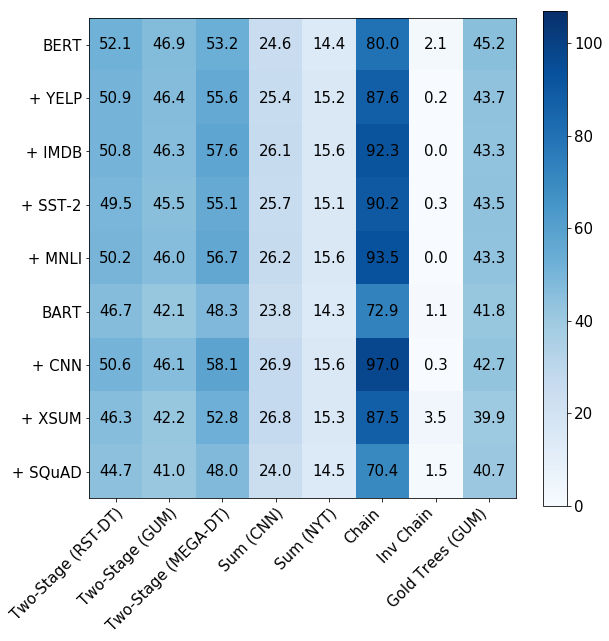}
    \caption{Detailed PLM discourse constituency (left) and dependency (right) structure overlap with baselines and gold trees according to the original parseval and UAS metrics.}
    \label{fig:detailed_intersection_baselines}
\end{figure*}
\begin{figure*}[h]
    \centering
    \includegraphics[width=.46\textwidth]{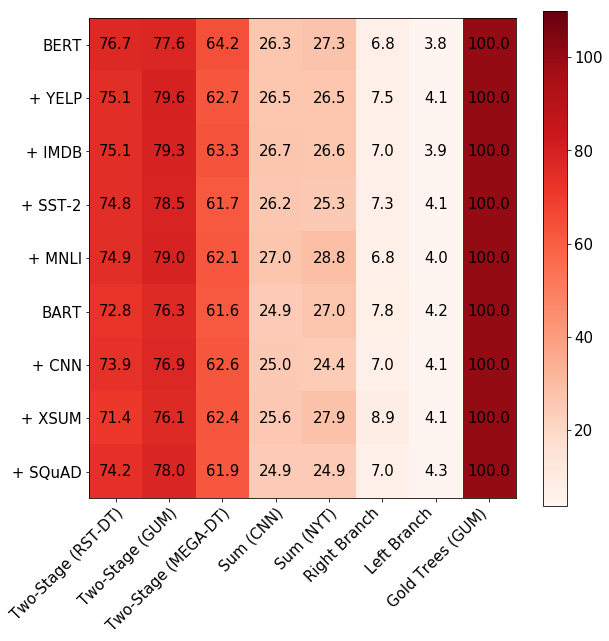}
    \includegraphics[width=.46\textwidth]{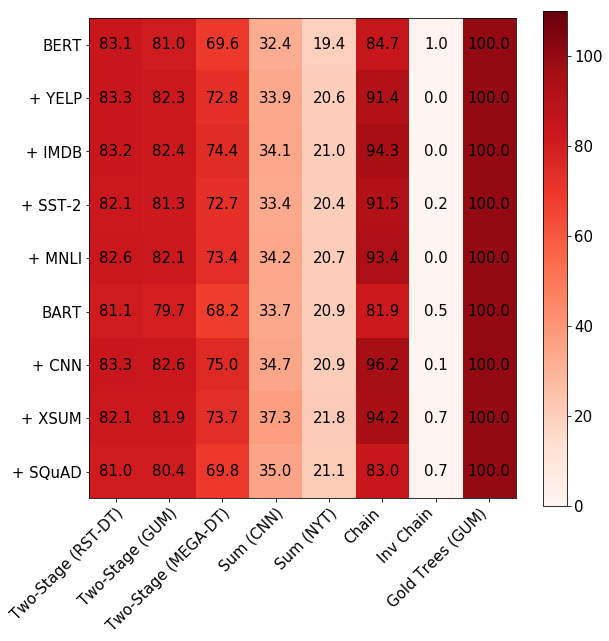}
    \caption{Detailed PLM discourse constituency (left) and dependency (right) structure performance of intersection with gold trees (e.g., BERT $\cap$ Gold Trees $\leftrightarrow$ Two-Stage (RST-DT) $\cap$ Gold Trees) according to the original parseval and UAS metrics.}
    \label{fig:detailed_intersection_baselines_with_gold}
\end{figure*}

\newpage
\section{Intra- and Inter-Model Self-Attention Comparison}
\label{app:self_attn_comp}

\begin{figure*}[h]
  \subfigure[BERT constituency tree similarity on GUM]{
	\begin{minipage}[c][1\width]{
	   0.45\textwidth}
	    \centering
	    \stackunder[3pt]{
        \includegraphics[width=.45\linewidth]{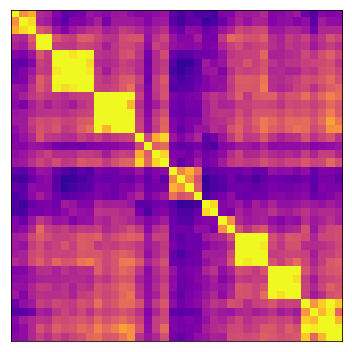}
        \includegraphics[width=.45\linewidth]{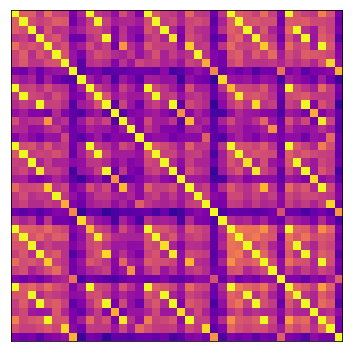}}{\footnotesize Heatmaps sorted by heads (left) and models (right)}\\
        \bigskip
        \stackunder[3pt]{
        \begin{tikzpicture}
          \begin{axis}
            [
            width=.8\linewidth,
            height=3.6cm,
            ytick={1,2,3,4},
            yticklabels={=Head*, $\neq$Head ~~, =Model ~~, $\neq$Model ~~},
            ticklabel style = {font=\small}
            ]
            \addplot+[
            boxplot prepared={
              median=0.670721,
              upper quartile=0.702731,
              lower quartile=0.617256,
              upper whisker=0.8210,
              lower whisker=0.490133
            },
            ] coordinates {};
            \addplot+[
            boxplot prepared={
              median=0.431849,
              upper quartile=0.475447,
              lower quartile=0.391005,
              upper whisker=0.596145,
              lower whisker=0.285452
            },
            ] coordinates {};
            \addplot+[
            boxplot prepared={
              median=0.435521,
              upper quartile=0.481413,
              lower quartile=0.397889,
              upper whisker=0.596145,
              lower whisker=0.31895
            },
            ] coordinates {};
            \addplot+[
            boxplot prepared={
              median=0.436898,
              upper quartile=0.481872,
              lower quartile=0.391923,
              upper whisker=0.821,
              lower whisker=0.285452
            },
            ] coordinates {};
          \end{axis}
        \end{tikzpicture}}{\footnotesize}
	\end{minipage}}
 \hfill 	
  \subfigure[BERT dependency tree similarity on GUM]{
	\begin{minipage}[c][1\width]{
	   0.45\textwidth}
	   \centering
	   \stackunder[3pt]{
        \includegraphics[width=.45\linewidth]{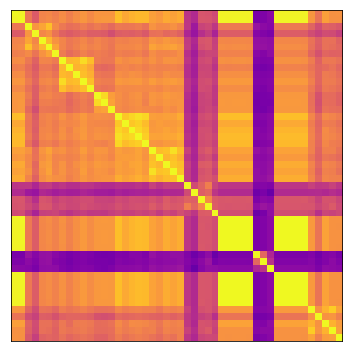}
        \includegraphics[width=.45\linewidth]{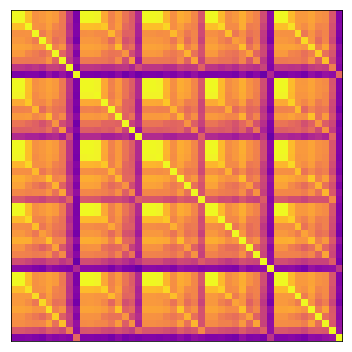}}{\footnotesize Heatmaps sorted by heads (left) and models (right)}\\
        \bigskip
        \stackunder[3pt]{
        \begin{tikzpicture}
          \begin{axis}
            [
            width=.8\linewidth,
            height=3.6cm,
            ytick={1,2,3,4},
            yticklabels={=Head*, $\neq$Head ~~, =Model ~~, $\neq$Model ~~},
            ticklabel style = {font=\small}
            ]
            \addplot+[
            boxplot prepared={
              median=0.949068,
              upper quartile=0.979764,
              lower quartile=0.910641,
              upper whisker=1.0,
              lower whisker=0.78126
            },
            ] coordinates {};
            \addplot+[
            boxplot prepared={
              median=0.888131,
              upper quartile=0.914052,
              lower quartile=0.826739,
              upper whisker=1.0,
              lower whisker=0.711687
            },
            ] coordinates {};
            \addplot+[
            boxplot prepared={
              median=0.887676,
              upper quartile=0.914507,
              lower quartile=0.831742,
              upper whisker=1.0,
              lower whisker=0.723
            },
            ] coordinates {};
            \addplot+[
            boxplot prepared={
              median=0.890859,
              upper quartile=0.915416,
              lower quartile=0.827649,
              upper whisker=1.0,
              lower whisker=0.71168
            },
            ] coordinates {};
          \end{axis}
        \end{tikzpicture}}{\footnotesize}
	\end{minipage}}
  \subfigure[BART constituency tree similarity on GUM]{
	\begin{minipage}[c][1\width]{
	   0.45\textwidth}
	    \centering
	    \stackunder[3pt]{
        \includegraphics[width=.45\linewidth]{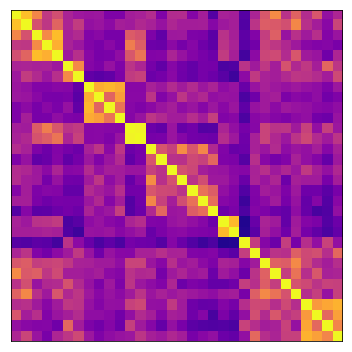}
        \includegraphics[width=.45\linewidth]{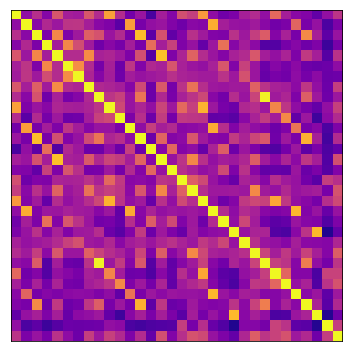}}{\footnotesize Heatmaps sorted by heads (left) and models (right)}\\
        \bigskip
        \stackunder[3pt]{
        \begin{tikzpicture}
      \begin{axis}
        [
        width=.8\linewidth,
        height=3.6cm,
        ytick={1,2,3,4},
        yticklabels={=Head*, $\neq$Head ~~, =Model*, $\neq$Model ~~},
        ticklabel style = {font=\small}
        ]
        \addplot+[
        boxplot prepared={
          median=0.571822,
          upper quartile=0.609454,
          lower quartile=0.526847,
          upper whisker=0.67462,
          lower whisker=0.478
        },
        ] coordinates {};
        \addplot+[
        boxplot prepared={
          median=0.418541,
          upper quartile=0.444699,
          lower quartile=0.390087,
          upper whisker=0.576870,
          lower whisker=0.29968
        },
        ] coordinates {};
        \addplot+[
        boxplot prepared={
          median=0.441028,
          upper quartile=0.469826,
          lower quartile=0.411771,
          upper whisker=0.57687,
          lower whisker=0.3162
        },
        ] coordinates {};
        \addplot+[
        boxplot prepared={
          median=0.414410,
          upper quartile=0.442061,
          lower quartile=0.388710,
          upper whisker=0.67462,
          lower whisker=0.2996788
        },
        ] coordinates {};
      \end{axis}
    \end{tikzpicture}}{\footnotesize}
	\end{minipage}}
 \hfill 	
  \subfigure[BART dependency tree similarity on GUM]{
	\begin{minipage}[c][1\width]{
	   0.45\textwidth}
	   \centering
	   \stackunder[3pt]{
        \includegraphics[width=.45\linewidth]{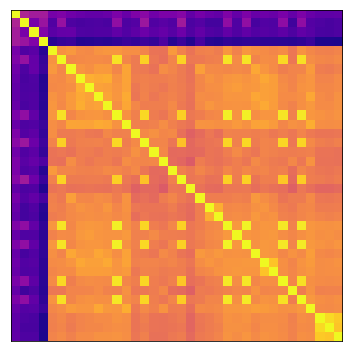}
        \includegraphics[width=.45\linewidth]{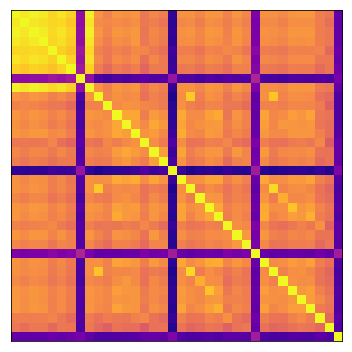}}{\footnotesize Heatmaps sorted by heads (left) and models (right)}\\
        \bigskip
        \stackunder[3pt]{
        \begin{tikzpicture}
      \begin{axis}
        [
        width=.8\linewidth,
        height=3.6cm,
        ytick={1,2,3,4},
        yticklabels={=Head*, $\neq$Head ~~, =Model*, $\neq$Model ~~},
        ticklabel style = {font=\small}
        ]
        \addplot+[
        boxplot prepared={
          median=0.897681,
          upper quartile=0.915985,
          lower quartile=0.880400,
          upper whisker=0.9736,
          lower whisker=0.72487
        },
        ] coordinates {};
        \addplot+[
        boxplot prepared={
          median=0.887221,
          upper quartile=0.903138,
          lower quartile=0.860960,
          upper whisker=0.99545,
          lower whisker=0.6489
        },
        ] coordinates {};
        \addplot+[
        boxplot prepared={
          median=0.899727,
          upper quartile=0.926671,
          lower quartile=0.870623,
          upper whisker=0.99545,
          lower whisker=0.64893
        },
        ] coordinates {};
        \addplot+[
        boxplot prepared={
          median=0.886085,
          upper quartile=0.900978,
          lower quartile=0.859936,
          upper whisker=0.9736,
          lower whisker=0.65029
        },
        ] coordinates {};
      \end{axis}
    \end{tikzpicture}}{\footnotesize}
	\end{minipage}}
\caption{Top: Visual analysis of sorted heatmaps. Yellow=high score, purple=low score.\\
Bottom: Aggregated similarity of same heads, same models, different heads and different models. *=Head/=Model significantly better than $\neq$Head/$\neq$Model performance with p-value $<0.05$.}
\label{fig:bart_sim_detailed}
\end{figure*}

\end{document}